\documentclass[opre,nonblindrev]{informs3} 

\DoubleSpacedXI 

\usepackage{natbib}
\bibpunct[, ]{(}{)}{,}{a}{}{,}%
\def\bibfont{\small}%
\usepackage{amsmath}
\usepackage{amssymb}
\usepackage{color}
\usepackage{graphicx,psfrag,epsf}
\usepackage{enumerate}
\usepackage{subfigure}
\usepackage{authblk}
\usepackage{parskip}
\usepackage{xcolor}
\usepackage{multirow}
\usepackage[linesnumbered,ruled,vlined]{algorithm2e}

\newcommand{\M}[1]{\boldsymbol{#1}}  
\newcommand{\V}[1]{\boldsymbol{#1}}  

\renewcommand{\vec}[0]{\textrm{vec}}

\allowdisplaybreaks[1]

\SetCommentSty{mycommfont}

\SetKwInput{KwInput}{Input}                
\SetKwInput{KwOutput}{Output}              
\SetKwInput{KwInit}{Initialization}

\TheoremsNumberedThrough     
\ECRepeatTheorems

\EquationsNumberedThrough    

\begin{document}
	
	
	\RUNAUTHOR{Park, Noh, Srivastava}
	
	\RUNTITLE{Data Science for Motion and Time Analysis}
	
	\TITLE{Data Science for Motion and Time Analysis with Modern Motion Sensor Data}
	
	\ARTICLEAUTHORS{%
		\AUTHOR{Chiwoo Park}
		\AFF{Department of Industrial and Manufacturing Engineering, Florida State University, Tallahassee, FL 32306, \EMAIL{cpark5@fsu.edu}} 
		\AUTHOR{Sang Do Noh}
		\AFF{Department of Systems Management Engineering, Sungkyunkwan University, Suwon, South Korea, \EMAIL{sdnoh@skku.edu}}
		
		\AUTHOR{Anuj Srivastava}
		\AFF{Department of Statistics, Florida State University, Tallahassee, FL 32306, \EMAIL{asrivastava@fsu.edu}}
	} 
	
	\ABSTRACT{%
		The {\it motion-and-time} analysis has been a popular research topic in operations research, especially for analyzing work performances in manufacturing and service operations. It is regaining attention as continuous improvement tools for lean manufacturing and smart factory. This paper develops a framework for data-driven analysis of work motions and studies their correlations to work speeds or execution rates, using data collected from modern motion sensors.  The past analyses largely relied on manual steps involving time-consuming stop-watching and video-taping, followed by manual data analysis. While modern sensing devices have automated the collection of motion data, the motion analytics that transform the new data into knowledge are largely underdeveloped. Unsolved technical questions include: How the motion and time information can be extracted from the motion sensor data, how work motions and execution rates are statistically modeled and compared, and what are the statistical correlations of motions to the rates? In this paper, we develop a novel mathematical framework for motion and time analysis with motion sensor data, by defining new mathematical representation spaces of human motions and execution rates and by developing statistical tools on these new spaces. This methodological research is demonstrated using five use cases applied to manufacturing motion data. 
	}%
	
	
	
	\maketitle
	
	%
	
	
	\section{Introduction} \label{sec:intro}
	A {\it motion-and-time} is an operations research technique that analyzes work motions and corresponding work execution times to create standardized work practices and identify the best way to perform tasks \citep{barnes1949motion}. This problem area originated in 1950s from the efforts of Frederick Taylor and the Gilbreth family, the pioneers of scientific management and operations research. It is regaining attention in recent years as parts of lean manufacturing, healthcare operations research and other service operations that naturally involve many human tasks \citep{meyersmotion, barger2004health, singh2011growth, urgo2019human}. This problem can also be found in smart factory settings. While smart factories seek replacements of human workers with industrial robotics and AI, they still involve a large proportion of manual workers. Fully automated factories are very hard to realize, given the trend of mass-customization and corresponding diversity in production lines and limitations in enabling technologies. In such mixed environments of human and robotic operations, quantifying and analyzing the manual works at the same level as robotic operations automatically sensed by the Internet of Things (IoT) sensors are imperative to realize the smart factories \citep{ferrari2018motion}. 
	
	In the past, \textit{motion-and-time} study has been performed manually with the time-intensive stop-watching and video recording of human motions, followed by manual data analysis. This is inefficient to be performed on a regular basis. In the conventional analysis, a work of interest is divided into work elements or subtasks. \textit{Time study} involves measuring the times spent by different workers on different work elements, using a time-keeping device such as a stopwatch, and analyzing the time measurements to define a standard work execution rate or evaluate the relative performance of the workers with respect to the standard rate. In \textit{Motion study}, work motions are filmed with elapsed times to identify time-efficient work motions. Many steps of the motion and time study for defining, measuring and analyzing works are performed manually, taking up a significant amount of time and effort. With advances in high quality motion sensors, the measurement of human motions can be generated with little human intervention, in the form of time series of locations of multiple body parts. However, the analysis tools that convert this new data type into important findings in the time and motion study are largely under-developed. This gives us a major motivation and background to develop modern data science for the motion and time study for a more automated and quantitative analysis. 
	
	Automated analysis of human motions has been developed and explored in many scientific areas including computer vision, psychology, human-machine interaction and surveillance. Motion data are captured using conventional cameras \citep{borges2013video}, depth cameras \citep{yang2012recognizing}, or the modern motion capture technology \citep{puthenveetil2015computer}.  Human postures are extracted from motion data in the form of human silhouettes or human skeletons, illustrated in Fig. \ref{fig1}. The time-lapsed sequences of human postures are recorded and analyzed for various purposes. The existing studies has been heavily focused on only motions, particularly human action recognition inferring what humans do from motion data, or kinetic studies relating motions to human kinetics or dynamics, e.g., torques and forces. In the existing studies, the temporal variability in the motion data, such as the execution rates of motions and the total running time periods, are regarded as unwanted sources of variation that hamper the accurate analysis of human motions. Consequently, they are often discarded and not considered in the analysis. Here we are interested in analyzing both the spatial and temporal variabilities of motions and analyzing their correlations for the purpose of work evaluation and efficiency analysis. Section \ref{sec:review} reviews the relevant prior works on the spatial and temporal variabilities, and Section \ref{sec:contribution} will discuss our contributions and describe the organization of this paper.
	
	\subsection{Related Works} \label{sec:review}
	Human motion analysis has been studied heavily in computer vision with a primary emphasis on human motion recognition. The review in this section is focused on the prior research directly relevant to this paper: modeling spatial variability and temporal variability of human motions. A broader review can be found in recent review papers  \citep{aggarwal2011human, chaquet2013survey, wang2016survey}. 
	
	\subsubsection{Spatial variability of motion}
	As illustrated in Figure \ref{fig1}, a human is popularly abstracted as the locations of joints in a human body, which are referred to as skeleton data. The joint locations contain important information on the posture of a human but are also influenced by other uninteresting (or nuisance) factors such as the size, location and orientation of a human body. A human posture can be attained by normalizing out those nuisance factors from the skeleton data. Then a human motion is represented by a temporal evolution of human postures. Different representations of motions in the existing studies are distinguished by how human postures are defined and how the temporal evolution of human postures are modeled.
	
	Wang and the co-authors used the relative position of a joint to another joint, for every pair of the joints in skeleton data \citep{wang2012mining, wang2013learning}. By the nature, the relative positions are invariant to the location of a human. Some post processing steps are then performed to remove the body orientation information from the relative positions. To capture the temporal evolution of a human motion, the authors constructed a temporal Fourier pyramid, which consists of the fourier coefficients of the relative joint positions over multi-resolution temporal partitions. \citet{yang2014effective} take the spatial and temporal differences of the 3D joint locations and applied principal component analysis on the combined feature space. 
	
	Another group of the study divides a human skeleton into body parts (e.g. upper arm, lower arm, neck and so on) and the features are extracted from each of the body parts. The temporal trajectories of the features are used to model a human motion. The commonly used features are the joint angles between the neighboring body parts \citep{ofli2014sequence, ohn2013joint}. The temporal changes of the angles over a human motion are then analyzed. \citet{vemulapalli2014human} finds the rotation and translation of one body part to best align it to another body part, for every pair of body parts, and the transformation matrices defining the rotation and translation are used to represent a human posture.  
	
	\citet{xia2012view} used the 3D histogram of joint locations to describe human postures. The linear discriminant analysis is used to extract low-dimensional features of the histograms, which are used to identify a few distinct patterns in the histograms. The temporal evolution of human postures is modeled as a Markov chain of the pattern labels. 
	
	\citet{devanne2013space} used the 3D joint locations to represent a human posture, and a human motion is represented as a trajectory of the 3D joint location over time. The major issue with this approach is that the 3D joint locations can vary significantly depending on unwanted nuisances such as the orientation and scaling of human bodies. This issue was fixed by \citet{amor2015action}, which used Kendall's shape representation of the joint locations to remove some of the nuisance factors from the 3D joint data.
	
	\subsubsection{Temporal variability of motion}
	Human motions can be performed at different execution rates over different periods of time. This variability is referred to as the temporal variability of motions. There are very few research articles in the literature targeting the temporal variability. In the existing articles, the temporal variations among different human motions have been considered as a nuisance factor in human motion recognition and is often discarded.  \citet{veeraraghavan2009rate} first discussed how different execution rates of motions can affect negatively the accuracy of a human motion recognition. To fix this issue, \citet{veeraraghavan2009rate} applied dynamic time warping (DTW) to temporally align the trajectories of human silhouettes for removing the temporal variability in the trajectories. The same approach was used by \citet{abdelkader2011silhouette} and \citet{gritai2009matching} in different contexts. \citet{su2014statistical} proposed the transported square root velocity function (TSRVF) of a trajectory on a Riemannian manifold and developed the temporal alignment of trajectories based on the distance between the corresponding TSRVF features. \citet{amor2015action} applied these developments to temporally aligning the trajectories of the shapes of skeleton data. 
	
	\subsection{Major contribution and organization of the paper} \label{sec:contribution}
	The major contribution of this paper is to develop a new data science for evaluating the performance of human works with motion sensor data. Technical contributions are summarized into three bullet points:
	\begin{itemize}
		\item Postures invariant to location, body scale and body ratios: The existing human posture representations and corresponding motion representations are defined using features invariant to the location and orientation of a human body \citep{wang2012mining, vemulapalli2014human} or location, orientation and scaling \citep{amor2015action}. Spatial variations of skeletons due to different body scales are not considered properly. The size of a body part can be different from that of another part, and the ratios of the part sizes differ from person to person. The person-to-person variation of the body ratios would create unwanted variation in skeleton data.  We redefine a human posture as the features of skeleton data invariant to location, global body scale and ratios of body parts. In our motivating examples, the overall orientation of a human body provides important clues as to what the subject is doing, so we did not impose invariance to orientation. However, extending this work with the orientation invariance would be straightforward.
		\item Modeling and analysis of the spatial variability of motions: We define a new mathematical representation of the posture features. Due to the invariance to location, global body scale and body ratios, the space is a nonlinear Riemannian manifold, and not a Euclidean space. We impose a Riemannian metric on this space to enable comparisons of different posture features, to define a probability measure over the manifold and then to define a new statistical model on the spatial and temporal variation of motions. Relevant statistical inferences and estimations are also developed. 
		\item Modeling and analysis of the temporal variability of motions: Unlike existing approaches that focus only on motion analysis, we are interested in both the spatial and temporal variability of motions and how the spatial variability is related to the temporal variability. To this end, we develop a supervised dimension reduction to identify the spatial variations of motions most influential to the temporal variations. 
	\end{itemize}
	In Section \ref{sec:math}, we lay down necessary mathematical foundations, defining the notions of human postures, motions and rate of motions. We use a skeletonized representation of a human and define a human posture as the features of the skeleton which are not affected by the variations of body scales and ratios of body parts across people. We define a space of postures as a Riemannian manifold with the Riemannian metric, which forms a basis to compare different postures, define their variations and to capture their variability using probability distributions. A sequence of postures defines a human motion, and the elapsed times associated with the sequence define the execution rate of the motion. In Section \ref{sec:stat}, we define the probability models over the motion and rate spaces. Specifically, we develop a probability distribution on posture sequences as a joint distribution of postures with auto-regressive means and develop an approach of the supervised dimension reduction type to identify the major motion variations most influential to the execution rates. Section \ref{sec:app} shows five use cases of the methodological developments with manufacturing motion datasets. Finally, we conclude in Section \ref{sec:conc} with discussions of the methodological and practical implications of this work.  \\
	
	\section{Mathematical Framework} \label{sec:math}
	\begin{figure}[ht!]
		\centering
		\includegraphics[width=0.8\textwidth]{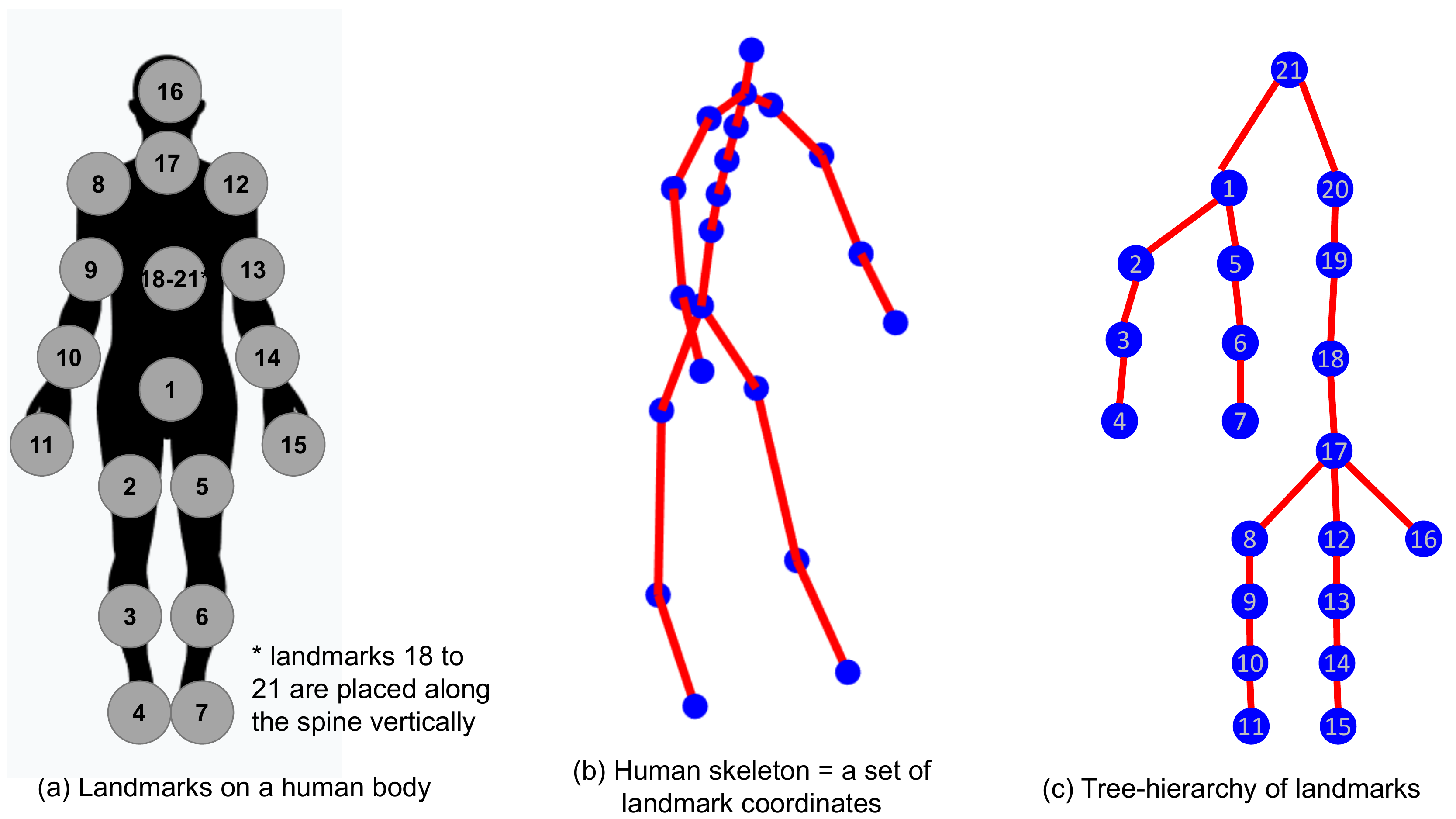}
		\caption{Human Skeleton. (a) shows the landmarks on a human body that represent where motion sensors are attached, (b) shows a skeleton of the landmarks, formed by connecting the landmarks with edges, and (c) shows the tree hierarchy of the landmarks in the skeleton.}
		\label{fig1}
	\end{figure}
	Human motions in a three dimensional (3D) space can be captured by different measurement devices such as conventional cameras from different angles, depth cameras and motion sensors. Human postures are measured periodically over an observation interval, and a sequence of the human postures determines a human motion.  As illustrated in Figure \ref{fig1}, a human posture is measured in the form of a set of landmark coordinates on the human's body, which is referred to as a human skeleton. Let $\M{X} \in \mathbb{R}^{n \times 3}$ denote a skeleton of $n$ landmark coordinates in a three-dimensional space. The skeleton data contains information about the body posture and also body's location, scale and orientation. We will discard the scale and location information to minimize the effect of the person-to-person variations in body scales and locations on our motion analysis. The orientation information is informative and will be retained, e.g., the orientation of a worker in a manufacturing cell gives some clues on what the worker is working on. In the existing works of the human motion data analysis \citep{amor2015action, gaglio2014human}, a centering and an overall scaling were applied to discard the scale and location from $\M{X}$, 
	\begin{equation*}
	\M{X}_s = \frac{\M{X} - \V{1}_n \V{c}^T}{s},
	\end{equation*}
	where $\V{c} \in \mathbb{R}^3$ represents the centroid of the $n$ landmarks, and $s$ is the mean distance of the $n$ landmarks to $\V{c}$. The global scaling is quite popular in landmark-based shape analysis due to \citet{kendall1989}. A major issue with such global scaling in human motion analysis is that people may have different body ratios, i.e., the size of one part of a body is different from that of another part, and the ratios of the body part sizes differ from person to person. The body ratios represent unwanted variables in our motion analysis and thus we want to discard the effect of the overall body size and body ratios. We define the \textit{posture} of a skeleton as the information remained after discarding the overall body size, body ratio and body location information from the skeleton. In Section \ref{sec:posture}, we define a mathematical representation of human postures, and in Section \ref{sec:motion}, we analyze motions and the rates of motions extracted from a sequence of postures. 
	
	\subsection{Posture Space of Skeletons} \label{sec:posture}
	First note that $n$ landmarks in the skeleton data form a tree-hierarchy, as illustrated in Figure \ref{fig1}-(c), where a landmark defines a tree node. Every node can be a root node. Just for convenience, we can order and label the landmarks with numbers and we choose the last landmark as a root. Let $\mathcal{T} = \{(i, j); j = parent(i), i = 1,\ldots, n-1\}$ represent the tree-hierarchy under the root. For $(i,j) \in \mathcal{T}$, the coordinate of the $i$th landmark relative to its parent $j$ can be defined as
	\begin{equation*}
	\V{\tilde{x}}_{i} = \V{x}_{i} - \V{x}_{j},
	\end{equation*}
	where $\V{x}_{i}  \in \mathbb{R}^3$ represent the coordinate of the $i$th landmark in $\M{X}$. The original data $\M{X}$ can be reconstructed using the originating root location $\V{x}_n$ and the relative coordinates $\M{\tilde{X}} = (\V{\tilde{x}}_{1}, \ldots, \V{\tilde{x}}_{n-1})$. To remove the location information from the skeleton, the root location will be discarded. The relative coordinates $\M{\tilde{X}}$ are invariant to the location of a human body, and only that information will be retained. The length of $\V{\tilde{x}}_{i}$, defined by the Euclidean norm $||\V{\tilde{x}}_{i}||$, quantifies the size of a body part connecting landmark $i$ and its parent. The relative ratio of the lengths defines the body ratio,
	\begin{equation*}
	\left\{\frac{||\V{\tilde{x}}_{i}||}{\sum_{k=1}^{n-1} ||\V{\tilde{x}}_{k}||}, i = 1,...,n-1 \right\},
	\end{equation*} 
	and the quantity $\sum_{k=1}^{n-1} ||\V{\tilde{x}}_{k}||$ represents the overall body size. We discard the overall size and body ratio by normalizing each $\V{\tilde{x}}_{i}$ to 
	\begin{equation*}
	\V{y}_{i} = \frac{\V{\tilde{x}}_{i}}{||\V{\tilde{x}}_{i}||}. 
	\end{equation*}
	Let $\M{Y} = (\V{y}_1, \V{y}_2, \ldots, \V{y}_{n-1})$ represents the body posture invariant to the location, body size and body ratio. 
	
	The posture is belongs to the product space,
	\begin{equation*}
	\mathcal{Y} = \smash[b]{\underbrace{\mathbb{S}^2 \times \ldots \times \mathbb{S}^2}_\text{$(n-1)$ times}},
	\end{equation*}
	where $\mathbb{S}^2$ is an unit sphere in $\mathbb{R}^3$, and $\times$ is the Cartesian product operator. The unit sphere is a Riemannian maniofold in $\mathbb{R}^3$, and $\mathcal{Y}$ is a product Riemannian manifold. The Riemannian metric on the product space can be induced from $\mathbb{S}^2$ and the product topology. For any two postures $\M{Y} \in \mathcal{Y}$ and $\M{Z} \in \mathcal{Y}$, the distance between them is defined as
	\begin{equation*}
	d_{\mathcal{Y}}(\M{Y}, \M{Z}) = \sum_{i=1}^{n-1} d_{\mathbb{S}^2}(\V{y}_i,  \V{z}_i),
	\end{equation*}
	where $\V{y}_i \in \mathbb{S}^2$ is the $i$th part of $\M{Y}$, $\V{z}_i \in \mathbb{S}^2$ is the $i$th part of $\M{Z}$, and $d_{\mathbb{S}^2}(\V{y}_i,  \V{z}_i) =  \cos^{-1}(\V{y}_i^T \V{z}_i)$ is the standard Riemannian distance on $\mathbb{S}^2$. Under this metric, the geodesic connecting two postures $\M{Y}$ and $\M{Z}$ is a vector of $(n-1)$ element-wise geodesics, 
	\begin{equation} \label{eq:geodesic}
	\V{g}_{\M{Y}, \M{Z}}(t) = (g_{\V{y}_1, \V{z}_1}(t), \ldots, g_{\V{y}_{n-1}, \V{z}_{n-1}}(t)),
	\end{equation}
	where $g_{\V{y}_i, \V{z}_i}(t)$ is the geodesic connecting $\V{y}_i$ and $\V{z}_i$ in $\mathbb{S}^2$. The form of the geodesic in $\mathbb{S}^2$ is well studied and can be found in literature \citep{srivastava2016functional}, which is given by
	\begin{equation*}
	g_{\V{y}_i, \V{z}_i}(t) = \frac{\sin(\theta_i - t\theta_i)}{\sin(\theta_i)} \V{y}_i + \frac{\sin(t\theta_i)}{\sin(\theta_i)}  \V{z}_i \mbox{ for } t \in [0, 1]
	\end{equation*}
	where $\theta_i = \cos^{-1}(\V{y}_i^T \V{z}_i)$. 
	
	\subsubsection{Tangent Space}
	In case of nonlinear Riemannian manifolds, it is more convenient to work on their tangent spaces, rather than working directly on the manifold, because a tangent space is an Euclidean space, and many statistical tools available on the Euclidean space can be applied. In this section, we define establish tangent spaces for the posture space $\mathcal{Y}$ and the corresponding exponential and inverse exponential maps that transform in between  $\mathcal{Y}$ and its tangent spaces. In particular, we are interested in defining a probability distribution on the tangent space for statistical analysis of postures.
	
	The tangent space of $\mathcal{Y}$ at a point $\M{Y}$ can be defined as a product of the element-wise tangent spaces, 
	\begin{equation*}
	T_{\M{Y}}(\mathcal{Y}) := T_{\V{y}_1}(\mathbb{S}^2) \times T_{\V{y}_2}(\mathbb{S}^2) \times \ldots \times T_{\V{y}_{n-1}}(\mathbb{S}^2),
	\end{equation*}
	where $T_{\V{y}_i}(\mathbb{S}^2)$ is the tangent space of $\mathbb{S}^2$ at $\V{y}_i$ given by
	\begin{equation*}
	T_{\V{y}_i}(\mathbb{S}^2) = \{\V{v} \in \mathbb{R}^3; \V{y}_i^T \V{v} = 0 \}.
	\end{equation*}
	
	The inverse exponential map $\exp^{-1}_{\M{Y}}(\M{Z})$ is a map projecting a point $\M{Z} \in \mathcal{Y}$ to the tangent space $T_{\M{Y}}(\mathcal{Y})$. Since the tangent space $T_{\M{Y}}(\mathcal{Y})$ is a direct product, the inverse exponential map can be defined element-wise. For $\V{y}_i, \V{z}_i \in \mathbb{S}^2$, the inverse exponential map for $\mathbb{S}^2$ is defined as
	\begin{equation*}
	\begin{split}
	\exp^{-1}_{\V{y}_i}(\V{z}_i) & = \frac{\theta_i}{\sin(\theta_i)} (\V{z}_i - \V{y}_i\cos(\theta_i)),
	\end{split}
	\end{equation*}
	where $\theta_i = \cos^{-1}(\V{y}_i^T \V{z}_i)$. The inverse exponential map $\exp^{-1}_{\M{Y}}(\M{Y})$ is thus defined as a vector of such maps,
	\begin{equation} \label{eq:iexp}
	\exp^{-1}_{\M{Y}}(\M{Z}) = (\exp^{-1}_{\V{y}_1}(\V{z}_1), \ldots, \exp^{-1}_{\V{y}_{n-1}}(\V{z}_{n-1})).
	\end{equation} 
	Please note that the tangent space $T_{\V{y}_i}(\mathbb{S}^2)$ is a two-dimensional vector space, spanned by a set of two basis vectors, in $\mathbb{R}^3$. The two orthonormal basis vectors can be obtained by applying the Gram-Schmidt process. Let $\V{\nu}_{\V{y}_i} \in T_{\V{y}_i}(\mathbb{S}^2)$ and $\V{\omega}_{\V{y}_i} \in T_{\V{y}_i}(\mathbb{S}^2)$ represent the two orthogonal basis vectors of $T_{\V{y}_i}(\mathbb{S}^2)$. The result of the inverse exponential map can be represented with the two corresponding basis coordinates $\V{c}_i = (c_{i1}, c_{i2})^T$,
	\begin{equation*}
	\exp^{-1}_{\V{y}_i}(\V{z}_i) = c_{i1} \V{\nu}_{\V{y}_i} + c_{i2} \V{\omega}_{\V{y}_i}.
	\end{equation*}
	Consequently, the coordinates for $\exp^{-1}_{\M{Y}}(\M{Z})$ are represented by a $2 \times (n-1)$ matrix,
	\begin{equation*}
	\M{C}_{\M{Z}|\M{Y}} = (\V{c}_1, \ldots ,\V{c}_{n-1}).
	\end{equation*}
	Likewise, the exponential map is also defined element-wise. Let $\V{F} \in T_{\M{Y}}(\mathcal{Y})$ denote a point in the tangent space, and let $\V{f}_i \in T_{\V{y}_i}(\mathbb{S}^2)$ represent a part of $\M{F}$ that belongs to $T_{\V{y}_i}(\mathbb{S}^2)$. The exponential map that transforms a point $\V{f}_i \in T_{\M{y}_i}(\mathbb{S}^2)$ back to a point in $\mathbb{S}^2$ is given by 
	\begin{equation*}
	\exp_{\V{y}_i}(\V{f}_i) = \cos(||\V{f}_i||) \V{y}_i+ \sin(||\V{f}_i||) \frac{\V{f}_i}{||\V{f}_i||}.
	\end{equation*}
	The full exponential map $\exp_{\M{Y}}(\mathcal{Y})$ is then defined as
	\begin{equation*}
	(\exp_{\V{y}_1}(\V{f}_1), \exp_{\V{y}_2}(\V{f}_2), \ldots, \exp_{\V{y}_{n-1}}(\V{f}_{n-1})).
	\end{equation*}
	
	\subsubsection{Parallel Transport on $\mathcal{Y}$}
	In geometry, a parallel transport on a manifold is the notion that implies translating a tangent space of the manifold along a differentiable curve to attain a new tangent space. In this section, we define a parallel transport in the posture manifold $\M{Y}$. Specifically, for two postures, $\M{Y} \in \mathcal{Y}$ and $\M{Z} \in \mathcal{Y}$, and for a tangent vector $\M{F} \in T_{\M{Y}}(\mathcal{Y})$, we defines a parallel transport of $\M{F}$ along the geodesic connecting $\M{Y}$ and $\M{Z}$. This is useful if one is interested in transporting a tangent vector from one tangent space to another. Since both $\mathcal{Y}$ and the tangent space $T_{\M{Y}}(\mathcal{Y})$ are the product spaces, and the parallel transport can be defined element-wise. The parallel transport of each $\V{f}_i$ follows the parallel transport in $\mathbb{S}^2$. The expression for parallel transport in $\mathbb{S}^2$ is available in an analytical form. For $\V{y}_i \in \mathbb{S}^2$, $\V{z}_i (\neq \V{y}_i) \in \mathbb{S}^2$ and $\V{f}_i \in T_{\V{y}_i}(\mathbb{S}^2)$, the parallel transport is given by 
	\begin{equation*}
	\M{v}_{\V{y}_i \rightarrow \V{z}_i} = \M{f}_i - \frac{2\V{f}_i^T \V{z}_i}{||\M{y}_i+\M{z}_i||^2} (\M{y}_i+\M{z}_i).
	\end{equation*}
	
	\subsubsection{Posture Distribution} \label{sec:posture_dist}
	Let $\M{Y} = (\V{y}_1, \ldots, \V{y}_{n-1}) \in \mathcal{Y}$ represent a posture, and let $\V{y} = (\V{y}_1^T, \ldots, \V{y}_{n-1}^T)^T$ denote the vectorization of the posture. We want to define a probability distribution $p(\V{y})$. Let $\V{\mu} \in \mathcal{Y}$ denote the unknown mean of the distribution, and let $\V{\mu}_i$ represent the part of the mean corresponding to $\V{y}_i$. We map $\M{y}$ to the tangent space $T_{\V{\mu}}(\mathcal{Y})$ using the inverse exponential map \eqref{eq:iexp}, resulting in the following standard coordinates of the tangent vector: 
	\begin{equation*}
	\M{C}_{\V{\mu}} = (\V{c}_{\V{y}_1|\V{\mu}_1}, \ldots, \V{c}_{\V{y}_{n-1}|\V{\mu}_{n-1}})
	\end{equation*}
	where $\V{c}_{\V{\mu}_i} = (\V{\nu}_{\V{\mu}_i}^T \exp^{-1}_{\V{\mu}_i}(\V{y}_i), \V{\omega}_{\V{\mu}_i}^T \exp^{-1}_{\V{\mu}_i}(\V{y}_i))^T$. Please note that 
	\begin{equation*}
	\begin{split}
	\V{c}_{\V{\mu}_i} &= \M{W}_{\V{\mu}_i}^T \exp^{-1}_{\V{\mu}_i}(\V{y}_i) \\
	&= \M{W}_{\V{\mu}_i}^T  \frac{\theta_{\V{\mu}_i}}{\sin(\theta_{\V{\mu}_i})} (\V{y}_i - \V{\mu}_i \cos(\theta_i)) \\
	& = \frac{\theta_{\V{\mu}_i}}{\sin(\theta_{\V{\mu}_i})}\M{W}_{\V{\mu}_i}^T \V{y}_i,
	\end{split}
	\end{equation*}
	where $\M{W}_{\V{\mu}_i} = (\V{\nu}_{\V{\mu}_i}, \V{\omega}_{\V{\mu}_i})$, and $\theta_{\V{\mu}_i} = \cos^{-1}(\V{\mu}_i^T \M{y}_i)$. The Euclidean norm of $\V{c}_{\V{y}_i|\V{\mu}_i}$ is bounded by
	\begin{equation*}
	\begin{split}
	||\V{c}_{\V{\mu}_i}|| & \le \left|\left|\frac{\theta_{\V{\mu}_i}}{\sin(\theta_{\V{\mu}_i})} \right|\right| \cdot ||\M{W}_{\V{\mu}_i}^T|| \cdot ||\V{y}_i||  \le \pi/2.
	\end{split}
	\end{equation*} 
	Let $\V{c}_{\V{\mu}} = \vec(\M{C}_{\M{y}|\V{\mu}}) \in \mathbb{R}^{2(n-1)}$ represent the vectorization of the standard tangent space coordinates. We impose a truncated multivariate normal distribution on $\V{c}_{\V{\mu}}$ with density
	\begin{equation} \label{eq:xdist}
	p(\V{c}_{\V{\mu}}|\V{\mu}, \M{K}) = a_K |\M{K}|^{-1/2} \exp\left\{ - \frac{1}{2}\V{c}_{\V{\mu}}^T \M{K}^{-1} \V{c}_{\V{\mu}}  \right\} q(\V{c}_{\V{\mu}}),
	\end{equation}
	where $\M{K}$ is a $2(n-1) \times 2(n-1)$ positive definite matrix, $a_K$ is the normalizing constant, and $q(\V{c}_{\V{\mu}}) = \prod_{i=1}^{n-1} I_{||\V{c}_{\V{\mu}_i}|| \le \pi/2}$ represents the support of the density. 
	Noting the relation,
	\begin{equation*}
	\V{c}_{\V{\mu}} = \M{W}^T_{\V{\mu}} \V{y},
	\end{equation*}
	where $\M{W}_{\V{\mu}}$ is a block diagonal matrix of $(n-1)$ blocks with $\frac{\theta_{\V{\mu}_i}}{\sin(\theta_{\V{\mu}_i})}\M{W}_{\V{\mu}_i}$ as its $i$ diagonal matrix, the wrapped truncated distribution of $\V{y}$ can be induced from $p(\V{c}_{\V{\mu}}|\V{\mu}, \M{K})$,
	\begin{equation} \label{eq:ydist}
	p(\V{y}|\V{\mu}, \M{K}) = a_K |\M{K}|^{-1/2} \exp\left\{ - \frac{1}{2}\V{y}^T \M{W}_{\V{\mu}} \M{K}^{-1} \M{W}_{\V{\mu}}^T \V{y}  \right\} \prod_{i=1}^{n-1} \left[\frac{\theta_{\V{\mu}_i}}{\sin(\theta_{\V{\mu}_i})}I_{||\theta_{\V{\mu}_i}||\le \pi/2} \right].
	\end{equation} 
	
	\subsection{Human Motion and Rate of Motion} \label{sec:motion}
\begin{figure}[ht!]
		\includegraphics[width=\textwidth]{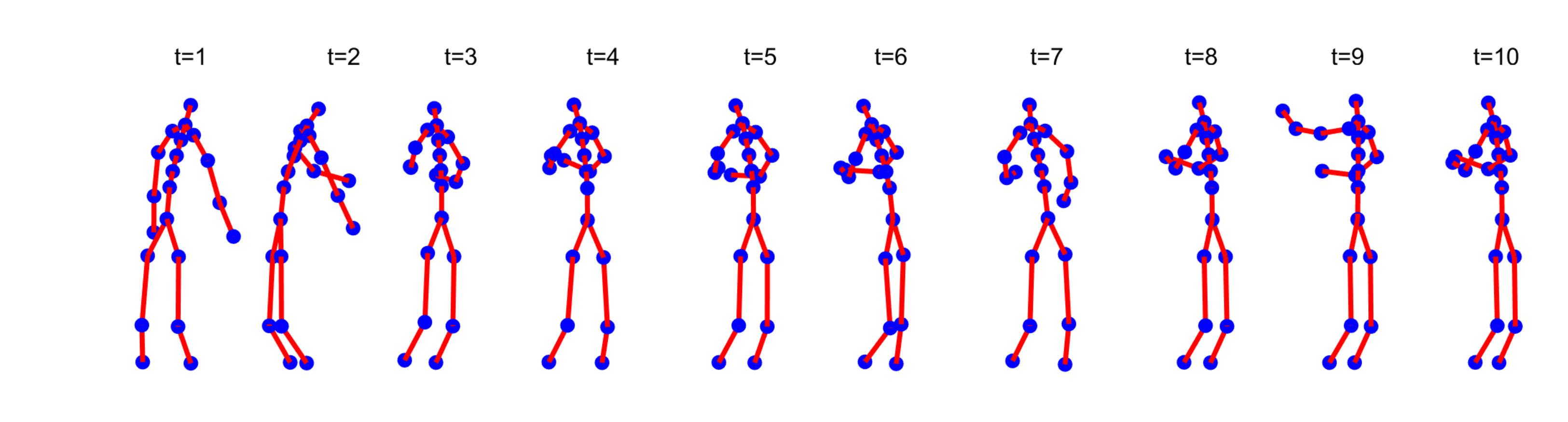}
		\caption{Human motion in a manufacturing station. A sequence of human postures defines a human motion, and the time labels over the postures defines the rate or speed of the motion.}
		\label{fig2}
	\end{figure}

	The human motion data is available to us as a sequence of human skeletons. After discarding the body scale and location information from each skeleton in the sequence, it can be converted into a sequence of postures. A posture sequence contains the information on the motion and the execution rate of the motion, where the \textit{motion} is defined as a series of postures with no time stamp, and the \textit{rate} is defined as the time stamp attached to the motion and it describes how fast the motion progresses. For example, Figure \ref{fig2} shows a sequence of postures. Depending on how to one assigns time indices to these postures, the sequence can be performed at different rates. For instance, the series can be time-stamped with $t=1$, $t=2$, $t=3$ and so on, or it can be time-stamped with $t=1$, $t=3$, $t=5$ and so on. The latter case is twice slower than the former case.  Our goal is to separate out the motion and the rate information from a posture sequence and to study their correlation.  
	
	\subsubsection{Motion and Motion Space}
	We start by defining the notion of a motion and its representation space. A posture sequence is a sequence of postures ordered by their appearance times. Suppose that there is a posture sequence observed for a physical time $t \in [0, U_{\alpha}]$. The time can be normalized to $[0, 1]$ by uniform scaling. A posture sequence at the normalized time scale is written as a smooth map $\alpha: [0, 1] \mapsto \mathcal{Y}$, where $\alpha(t)$ represents the human posture at a normalized time $t$. Let $\mathcal{A}$ denote a space of all such maps. 
	
	The time-stamping of a posture sequence $\alpha$ can be obtained by composing $\alpha$ with a strictly increasing function $\gamma: [0, 1] \mapsto [0, 1]$. The newly indexed sequence, $\alpha \circ \gamma$, also called a reparameterization of $\alpha$, is a posture sequence belonging to $\mathcal{A}$. The new sequence contains the exactly same postures as $\alpha$ but comes with a different time stamping; the posture $\alpha(\gamma(t))$ has time stamp $\gamma(t)$ in a posture sequence $\alpha$ but the same posture has time stamp $t$ in $\alpha \circ \gamma$. Therefore, one can generate differently time-stamped versions of $\alpha$ with different choices of $\gamma$. The motion of $\alpha$ is deemed invariant to the choice of the reparameterization. For a more formal definition, let $\mathcal{G}$ represent a set of all diffeomorphisms from $[0, 1]$ to $[0, 1]$,
	\begin{equation*}
	\mathcal{G} = \{\gamma: [0, 1] \mapsto [0,1]; \gamma \mbox{ is a diffeomorphism},\ \gamma(0) = 0, \gamma(1) = 1\}. 
	\end{equation*}
	By the definition of a diffeomorphism, $\gamma$ is differentiable and bijective, and its inverse is also differentiable. Since every continuous and bijective function is strictly monotonic, $\mathcal{G}$ is a subset of all strictly increasing and surjective functions from $[0, 1]$ to $[0, 1]$. In fact, $\mathcal{G}$ is dense in a space of all strictly increasing and surjective functions with respect to the $L^2$ norm. Therefore, we restrict the space of the reparameterizations to $\mathcal{G}$. A diffeomorphism $\gamma \in \mathcal{G}$ defines a group action on $\mathcal{A}$ in that if $\alpha(t) \in \mathcal{A}$,  then $(\alpha \circ \gamma)(t)$ is also in $\mathcal{A}$ for every $\gamma \in \mathcal{G}$. The group action does not change postures and their orders in the sequence but only changes the time-stamping of the postures in the sequence. A motion of $\alpha$ is equivalent to that of $\alpha \circ \gamma$ for $\gamma \in \mathcal{G}$. A motion is defined as an equivalence class of $\alpha$, 
	\begin{equation*}
	[\alpha] = \{\alpha \circ \gamma \in \mathcal{A}; \gamma \in \mathcal{G}\},
	\end{equation*}
	and a space of motions is the quotient space $\mathcal{A} / \mathcal{G}$. A proper metric to quantify differences between two motions $[\alpha_1]$ and $[\alpha_2]$ is then defined as 
	\begin{equation*}
	d_{\mathcal{A} / \mathcal{G}}([\alpha_1], [\alpha_2]) = \min_{\gamma_1, \gamma_2 \in \mathcal{G}} \int_{[0, 1]} d_{\mathcal{Y}}(\alpha_1 \circ \gamma_1, \alpha_2 \circ \gamma_2) dt. 
	\end{equation*}
	Numerically, calculating the distance involves optimizing $\gamma_1$ and $\gamma_2$ jointly, which can be complicated. 
	
	Alternatively, following \citet{su2014statistical}, we can transform $\alpha \in \mathcal{A}$ to the transported square-root vector field (TSRVF) with respect to a reference posture $\V{Y}_R \in \mathcal{Y}$,
	\begin{equation*}
	h_{\alpha} = \frac{\dot{\alpha}(t)_{\alpha(t) \rightarrow \V{Y}_R} }{ \sqrt{ ||\dot{\alpha}(t)||}},
	\end{equation*}
	where $\dot{\alpha}(t) \in T_{\alpha(t)}$ is the first order derivative of $\alpha(t)$ with respect to $t$, and $||\cdot||$ is the Euclidean norm on the tangent space $T_{\alpha(t)}(\mathcal{Y})$.  The derivative $\dot{\alpha}(t)$ belongs to the tangent space $T_{\alpha(t)}(\mathcal{Y})$, and $\dot{\alpha}(t)_{\alpha(t) \rightarrow \V{Y}_R}$ is the parallel transport of the tangent vector to the tangent space  $T_{\M{Y}_R}(\mathcal{Y})$. Therefore, the TSRVF $h_{\alpha}$ can be seen as a parameterized curve in $T_{\M{Y}_R}(\mathcal{Y})$. Let $\mathcal{H} = \{h_{\alpha}: \alpha \in \mathcal{A}\}$ represents the space of TSRVFs on $\mathcal{A}$. The distance between the TSRVFs of two posture sequences, $\alpha_1$ and $\alpha_2$, is defined as
	\begin{equation*}
	d_{\mathcal{H}}(h_{\alpha_1}, h_{\alpha_2}) = \int_{[0, 1]} ||h_{\alpha_1}(t) - h_{\alpha_2}(t)|| dt.
	\end{equation*}
	An useful property of the distance \citep{su2014statistical} is that for $\gamma \in \mathcal{G}$, 
	\begin{equation*}
	h_{\alpha \circ \gamma}(t) = h_{\alpha}(\gamma(t)) \sqrt{\dot{\gamma}(t)},
	\end{equation*}
	and for all $\gamma\in \mathcal{G}$
	\begin{equation*}
	d_{\mathcal{H}}(h_{\alpha_1 \circ \gamma}, h_{\alpha_2 \circ \gamma}) = d_{\mathcal{H}}(h_{\alpha_1}, h_{\alpha_2}).
	\end{equation*}
	Therefore, for any $\gamma_1 \in \mathcal{G}$ and $\gamma_2 \in \mathcal{G}$, there exists $\gamma \in \mathcal{G}$ satisfying
	\begin{equation*}
	\begin{split}
	d_{\mathcal{H}}(h_{\alpha_1 \circ \gamma}, h_{\alpha_2 \circ \gamma}) &= d_{\mathcal{H}}(h_{\alpha_1 \circ \gamma_1 \circ \gamma_2^{-1}}, h_{\alpha_2 \circ \gamma_2 \circ \gamma_2^{-1}})\\
	& = d_{\mathcal{H}}(h_{\alpha_1 \circ \gamma_1 \circ \gamma_2^{-1}}, h_{\alpha_2})\\
	& = d_{\mathcal{H}}(h_{\alpha_1 \circ \gamma}, h_{\alpha_2}).
	\end{split} 
	\end{equation*}
	Exploiting the result, we define the distance between two motions $[\alpha_1]$ and $[\alpha_2]$ as 
	\begin{equation*}
	\begin{split}
	d_{\mathcal{H} / \mathcal{G}}([\alpha_1], [\alpha_2]) &= \min_{\gamma_1, \gamma_2 \in \mathcal{G}} d_{\mathcal{H}}(h_{\alpha_1 \circ \gamma_1}, h_{\alpha_2 \circ \gamma_2}) \\
	& = \min_{\gamma \in \mathcal{G}} d_{\mathcal{H}}(h_{\alpha_1 \circ \gamma}, h_{\alpha_2}).
	\end{split} 
	\end{equation*}
	The distance $d_{\mathcal{H} / \mathcal{G}}$ is much easier to calculate than the distance $d_{\mathcal{A} / \mathcal{G}}$, using the dynamic programming. We use $d_{\mathcal{H} / \mathcal{G}}$ as the distance in our motion space $\mathcal{A}/ \mathcal{G}$. 
	
	\subsubsection{Rate Space} \label{sec:rate}
	The rate of a posture sequence can be defined only relatively with respect to a reference posture sequence. Let $\alpha \in \mathcal{A}$ represents a posture sequence of a worker following a standard operating procedure (SOP), and let $\alpha_R \in \mathcal{A}$ denote a reference posture sequence following the same manual. Please note that the posture sequences are defined on the normalized time horizon, $[0, 1]$, by dividing the physical time with the respective total running times, $U_{\alpha}$ and $U_{\alpha_R}$. If $U_{\alpha} = U_{\alpha_R}$, then after normalization, the relative speed of $\alpha$ with respect to the reference $\alpha_R$ is defined as the reparameterization that aligns $\alpha$ to $\alpha_R$ in time, according to:
	\begin{equation*}
	\gamma_{\alpha} = \argmin_{\gamma \in \mathcal{G}} d_{\mathcal{H}}(h_{\alpha \circ \gamma}, h_{\alpha_R}).
	\end{equation*}
	Note that the posture of $\alpha$ at time $\gamma_{\alpha}(t)$ corresponds to the posture of $\alpha_R$ at time $t$. If $\gamma_{\alpha}(t) > t$, the posture sequence $\alpha$ goes more slowly than the reference sequence at time $\gamma_{\alpha}(t)$. The first order derivative $\dot{\gamma}_{\alpha}(t)$ defines the instant speed of $\alpha$ at time $\gamma_{\alpha}(t)$. Since $\alpha$ runs the action corresponding to $\alpha_R(t)$,  $\dot{\gamma}_{\alpha}(t)$ can be interpreted as the speed of the action $\alpha_R(t)$ in the posture sequence $\alpha$. 
	
	When the total running times are different, the relative speed of $\alpha$ is defined as $(U_{\alpha} / U_{\alpha_R}) \gamma_{\alpha}$, a scaled version of $\gamma_{\alpha}$. Without loss of generality, we can set $U_{\alpha_R} = 1$, and the relative speed of $\alpha$ is defined as 
	\begin{equation*}
	\delta_{\alpha} = U_{\alpha} \gamma_{\alpha}.
	\end{equation*}
	Its first order derivative $\dot{\delta}_{\alpha}$ defines the relative instant speed of $\alpha$. Please note that $\delta_{\alpha_R}(t) = t$ is a simple identity map, and $\dot{\delta}_{\alpha_R}(t) = 1$ for every $t \in [0, 1]$. This implies that the instant speed of the reference sequence $\alpha_R(t)$ is constantly one, and $\dot{\delta}_{\alpha}$ defines the instant speed of $\alpha$ relative to the baseline. In other words, $\dot{\delta}_{\alpha} < 1$ implies that $\alpha$ is running slower than the reference sequence at time $\gamma_{\alpha}(t)$, and $\dot{\delta}_{\alpha} > 1$ implies that $\alpha$ is running faster at time $\gamma_{\alpha}(t)$. The relative speed $\dot{\delta}_{\alpha}$ is defined as the rate of motion $\alpha$.  
	
	The rate $\dot{\delta}_{\alpha}$ belongs to a class of all non-negative functions defined on $[0, 1]$,
	\begin{equation*}
	\mathcal{R} = \{ \dot{\delta}: [0, 1] \mapsto \mathbb{R}^+\}.
	\end{equation*}
	It is not a vector space, because a linear combination of the functions in $\mathcal{R}$ is not in $\mathcal{R}$. It is a convex cone. The log of the rate function belongs to a class of real-valued functions,
	\begin{equation*}
	\Gamma = \{ r: [0, 1] \mapsto \mathbb{R}\}.
	\end{equation*}
	It is a vector space. We can use the standard $L_2$ metric to define the distance between $r_1 ,r_2 \in \Gamma$, 
	\begin{equation*}
	d_{\Gamma}(r_1, r_2) = \left(\int_0^1 (r_1-r_2)^2 dt\right)^{1/2}.
	\end{equation*}
	
	\section{Statistical Analysis of Motion and Time} \label{sec:stat}
	Suppose that we observe movements of $M$ different workers, each following a standard operating procedure (SOP), resulting in $M$ sequences of skeletons. We want to perform the statistical analysis of work motions and the rates of the motions with the observations. Let $\alpha_m \in \mathcal{A}$ denote the posture sequence of the $m$th observed skeleton sequence, and let $\gamma_m \in \mathcal{G}$ represent the relative execution rate of $\alpha_m$ with respect to a reference posture sequence $\alpha_R \in \mathcal{A}$, 
	\begin{equation*}
	\gamma_m = \argmin_{\gamma \in \mathcal{G}} d_{\mathcal{H}}(h_{\alpha_m \circ \gamma}, h_{\alpha_R}),
	\end{equation*}
	The rate-normalized posture sequence of $\alpha$ can be defined as $\tilde{\alpha}_m = \alpha_m \circ \gamma_m$. We let the rate-normalized posture sequence represent the motion of $\alpha$, and the execution rate of the motion is defined as $r_m(t) = \log(\dot{\delta}_{\alpha_m}(t))$ as described in Section \ref{sec:rate}. 
	
	The rate-normalized posture sequences and the log rates are sampled at the $L$ discrete time points by the local kernel regression that will be introduced in Section \ref{sec:kernel_reg}, 
	\begin{equation*}
	\mathcal{D} = \{(\tilde{\alpha}_m(t_l), r_{m}(t_l)); m = 1,\ldots, M, l = 1, \ldots, L \}.
	\end{equation*} 
	We use the probability distribution of postures (defined in Section \ref{sec:posture_dist}) to impose a probability distribution on motions, as a joint distribution of the $L$ discrete postures, in Section \ref{sec:pos_dist}. We define a probability distribution of the execution rates using Gaussian process in Section \ref{sec:rate_dist}. Based on the distribution models, we formulate a sliced inverse regression to analyze the motions and execution rates jointly to identify the motion features that influence the work rates most; this is described in Section \ref{sec:edr}.
	
	\subsection{Local kernel regression for interpolating postures} \label{sec:kernel_reg}
	Let $u$ denote a covariate belonging to $\mathbb{R}$ and let $\M{Y}(u)$ denote the posture associated with the covariate. Given $S$ pairs of observations, $\mathcal{D}_Y = \{(u_s, \M{Y}_s) \in \mathbb{R}  \times \mathcal{Y}; s = 1, \ldots, S \}$, we would like to estimate $\M{Y}(u)$ at a test point $u \in \mathbb{R}$. We adopt and extend the conventional local kernel regression defined on the Euclidean space for obtaining the local kernel estimate of $\M{Y}(u)$. Similar to the conventional kernel estimate, we do not assume an explicit distribution of $\M{Y}(u)$ and only assume that the observed posture $\M{Y}_s$ is an independent sample with its expectation $E[\M{Y}_s] = \M{Y}(u_s)$ and random variation around the expectation, $E[d_{\mathcal{Y}}(\M{Y}_s, \M{Y}(u_s))^2] = \sigma_{y}^2$. A local kernel estimate $\M{Y}(u)$ is defined as $\hat{\M{Y}}(u) \in \mathcal{Y}$ that minimizes the locally weighted approximation error,  
	\begin{equation} \label{eq:local_kernel}
	\begin{split}
	\hat{\M{Y}}(u) & = \arg\min_{\M{Y} \in \mathcal{Y}} \sum_{s=1}^S d_{\mathcal{Y}}(\M{Y}, \M{Y}_s)^2 K_h(|u-u_s|),
	\end{split}
	\end{equation}
	where $K_h$ is a kernel function with bandwidth $h$ satisfying $\lim_{N \rightarrow \infty} h = 0$. Let $\omega_s = K_h(|u-u_s|)$ denote the kernel weight. The objective function in \eqref{eq:local_kernel} basically quantifies the sum of the squared geodesic distances to the observations, $d_{\mathcal{Y}}(\M{Y}, \M{Y}_s)^2$, weighted by the kernel weight. The problem \eqref{eq:local_kernel} can be rewritten as
	\begin{equation} \label{eq:local_kernel2}
	\begin{split}
	\hat{\M{Y}}(u) & = \arg\min_{\M{Y} \in \mathcal{Y}} \sum_{s=1}^S  \omega_s d_{\mathcal{Y}}(\M{Y}, \M{Y}_s)^2.
	\end{split}
	\end{equation}
	A gradient-based approach for solving the problem of this type has been studied in literature \citep{kurtek2012statistical}. Following the approach, the solution can be achieved by the  iterative steps described in Algorithm \ref{alg:localkernel}. 
	
	This algorithm will be applied in different parts of this paper. When the time $t$ corresponds to the covariate $u$ and the associated posture is the observation of a chosen posture sequence $\tilde{\alpha}_m$ at time $t$, this algorithm will find the local kernel estimates of $\tilde{\alpha}_m$ at regularly spaced time points $t_1, t_2, \ldots, t_L$, given the observation of $\tilde{\alpha}_m$ at irregularly spaced time points. This application is used for generating $\mathcal{D}$. When the execution rate $r$ corresponds to the covariate $u$, the algorithm will give the local kernel estimate of a posture that is associated with the execution rate, which will be applied to estimate the conditional expectation $E[\tilde{\alpha}_m(t_l)| r_{m,l}]$ in Section \ref{sec:edr}.
	
	\begin{algorithm}[ht!]
		\DontPrintSemicolon
		
		\KwInput{observations $\mathcal{D}_Y$, test point $u$}
		\KwOutput{local kernel estimate $\hat{\M{Y}}(u)$}
		{Find $s^* = \argmin_{s = 1,\ldots, S} |u_s - u|$ and set the initial estimate $\hat{\M{Y}}(u)$ to $Y_{s^*}$. }\\
		\For{$j = 1:J$}
		{
			Compute $\V{F}_{s} = \exp^{-1}_{\hat{\M{Y}}(u)}(\M{Y}_s)$ for $s = 1,\ldots, S$ \\
			Compute $$\bar{\V{F}} = \frac{\sum_{s=1}^S w_s \V{F}_{m,l}}{\sum_{s=1}^S w_s}.$$ \\
			\If{$||\bar{\V{F}}||_F \ge \xi$}
			{
				Update $\hat{\M{Y}}(u) = \exp_{\hat{\M{Y}}(u)}(\bar{\V{F}})$.
			}
			\Else
			{
				Stop the iteration.  
			}
		}
		\caption{Local kernel estimation of a posture} \label{alg:localkernel}
	\end{algorithm}
	
	\subsection{Probability distribution of motion} \label{sec:pos_dist}
	For each $l \in \{1,...,L\}$, $\tilde{\alpha}_m(t_l) \in \mathcal{Y}$ is a $3 \times (n-1)$ matrix with $(n-1)$ coordinates in $\mathbb{S}^2$. Let $\V{y}^{(m,l)}$ denote the vectorization of $\tilde{\alpha}_m(t_l)$. Following the wrapped normal model \eqref{eq:ydist}, we model the probability distribution of $\V{y}^{(m,l)}$ as 
	\begin{equation*}
	\begin{split}
	p(\V{y}^{(m,l)}|\V{\mu}^{(l)}, \M{K}_l) =& a_l |\M{K}_l|^{-1/2} \exp\left\{ - \frac{1}{2}(\V{y}^{(m,l)})^T \M{W}_{\V{\mu}^{(l)}} \M{K}_l^{-1} \M{W}_{\V{\mu}^{(l)}}^T \V{y}^{(m,l)}  \right\} \\
	& \qquad \prod_{i=1}^{n-1} \left[\frac{\theta_{\V{\mu}^{(l)}_i}}{\sin(\theta_{\V{\mu}^{(l)}_i})}  I_{\left|\left|\theta_{\V{\mu}^{(l)}_i}\right|\right|\le \pi/2} \right]
	\end{split}
	\end{equation*}
	where $\V{\mu}^{(l)} \in \mathcal{Y}$ is the mean, $\M{K}_l$ is the covariance parameter, and $a_l$ is the normalizing constant. We can also take the tangent coordinate $\V{c}^{(m, l)} = \M{W}_{\V{\mu}^{(l)}}^T \V{y}^{(m,l)}$ and use the truncated normal density on the tangent coordinate,
	\begin{equation} \label{eq:xlike}
	p(\V{c}^{(m, l)}|\V{\mu}^{(l)}, \M{K}_l) = a_l |\M{K}_l|^{-1/2} \exp\left\{ - \frac{1}{2}(\V{c}^{(m, l)})^T \M{K}^{-1}_l \V{c}^{(m, l)}   \right\} q(\V{c}^{(m, l)}).
	\end{equation}
	
	\subsubsection{Model with uncorrelated means}
	For simplicity, one can assume the mutual independence among the mean parameters and the covariance parameters. For such case, $\V{\mu}^{(l)}$ and $\M{K}_l$ can be estimated independently for each $l = 1,\ldots, L$. We take the maximum likelihood estimates with the relevant data, $\mathcal{D}_l = \{\V{y}^{(m,l)}; m = 1,\ldots, M\}$. The negative log likelihood is
	\begin{equation*}
	\begin{split}
	\mathcal{L}(\V{\mu}^{(l)}, \V{K}_l) =& \sum_{m=1}^M -2\log p(\V{c}^{(m,l)}|\V{\mu}^{(l)}, \M{K}_l) \\
	=&  M \log|\M{K}_l| + \sum_{m=1}^M  (\V{c}^{(m,l)})^T \M{K}_l^{-1} \V{c}^{(m,l)}.
	\end{split}
	\end{equation*}
	Algorithm \ref{alg:MLE} presents the iterative algorithm to obtain the maximum likelihood estimates (MLE) of $\V{\mu}^{(l)}$ and $\M{K}_l$. Please find the detailed derivation of the iterative algorithm in e-companion \ref{EC1}.

	\begin{algorithm}[ht!]
		\DontPrintSemicolon
		
		\KwInput{data $\mathcal{D}_l$, maximum number of iteration $J_{max}$, tolerance $\xi_1 > 0$ }
		\KwOutput{MLE estimates of $\V{\mu}^{(l)}, \M{K}_l$}
		{Initialize $\V{\mu}^{(l,0)}$ as one of $\V{y}^{(m,l)}$'s and set $\M{K}_{l,0} = \M{I}$.}\\
		\For{$j = 1:J_{max}$}
		{
			Compute $\M{W}_{\V{\mu}^{(l,j)}}^T \V{\mu}^{(l, j+1)} = \frac{1}{M} \sum_{m=1}^M \V{c}^{(m,l,j)}$; find details in \eqref{eq:mu_mle}. \\
			Update $\V{\mu}^{(l, j+1)} =  \exp_{\V{\mu}^{(l,j)}}(\tilde{\M{W}}_{\V{\mu}^{(l,j)}}\M{W}_{\V{\mu}^{(l,j)}}^T \V{\mu}^{(l, j+1)})$. \\
			
			\If{$||\M{W}_{\V{\mu}^{(l,j)}}^T \V{\mu}^{(l, j+1)}||_F \le \xi_1$}
			{
				Stop the iteration with the final iteration $j^* = j+1$. \\
			}
		}
		Set $\V{\hat{\mu}}^{(l)} = \V{\mu}^{(l, j^*)}$ and set $\V{\hat{K}}_{l} = \frac{1}{M} \sum_{m=1}^M \V{c}^{(l,m,j^*)} (\V{c}^{(l,m,j^*)})^T$; find details in \eqref{eq:K_mle}.
		\caption{MLE of the distribution parameters} \label{alg:MLE}
	\end{algorithm}
	
	\subsubsection{Model with autocorrelated means}
	We assume that the means are temporally correlated through the following autocorrelation model: for $l = 1,\ldots, L$,  
	\begin{equation*}
	\begin{split}
	p(\V{\mu}^{(l)}|\V{\mu}^{(l-1)}, \lambda_0^2 \M{I}) =& a_0 \lambda_0^{1-n} \exp\left\{ - \frac{1}{2\lambda_0^2}(\V{\mu}^{(l)})^T \M{W}_{\V{\mu}^{(l-1)}} \M{W}_{\V{\mu}^{(l-1)}}^T \V{\mu}^{(l)}  \right\} \\
	& \qquad \prod_{i=1}^{n-1} \left[\frac{\theta_{\V{\mu}^{(l-1)}_i}}{\sin(\theta_{\V{\mu}^{(l-1)}_i})}  I_{\left|\left|\theta_{\V{\mu}^{(l-1)}_i}\right|\right|\le \pi/2} \right],
	\end{split}
	\end{equation*}
	where $\lambda_0^2$ is the overall variance parameter, and $\V{\mu}^{(0)} \in \mathcal{Y}$ is the mean of $\V{\mu}^{(1)}$. Given the prior model, we seek for the maximum a posteriori probability (MAP) estimates of $\{\V{\mu}^{(l)}, \M{K}_l, l = 1,..,L \}$. The complete coordinate descent algorithm to achieve the MAP estimates is described in Algorithm \ref{alg:MAP}. Please find the detailed derivation of the estimation algorithm in e-companion \ref{EC2}. 
	
	\begin{algorithm}[ht!]
		\DontPrintSemicolon
		
		\KwInput{data $\mathcal{D}$, hyperparameters $\lambda_0^2$, $\V{\mu}_0$, $\M{K}_0$ }
		\KwOutput{MAP estimates of $\{\V{\mu}^{(l)}, \M{K}_l, l = 1,..,L \}$}
		{Find the initial estimate of $\{\V{\mu}^{(l)}, \M{K}_l, l = 1,..,L \}$ using Algorithm \ref{alg:MLE}, which are denoted by $$\{\V{\mu}^{(l,1)}, \M{K}_{l,1}, l = 1,..,L \}.$$}\\
		\For{$j = 1:J$}
		{
			\For{$l = 1:L$}
			{
				Compute $\M{W}_{\V{\mu}^{(l,j)}}^T \V{\mu}^{(l, j+1)}$ using \eqref{eq:mu}. \\
				Update $\V{\mu}^{(l, j+1)} =  \exp_{\V{\mu}^{(l,j)}}(\tilde{\M{W}}_{\V{\mu}^{(l,j)}}\M{W}_{\V{\mu}^{(l,j)}}^T \V{\mu}^{(l, j+1)})$. \\
				Update $\M{K}_{l, j+1}$ using \eqref{eq:K} 
			}
			
			\If{$\sum_{l=1}^L ||\M{W}_{\V{\mu}^{(l,j)}}^T \V{\mu}^{(l, j+1)}||_F \le \xi_2$}
			{
				Stop the iteration. 
			}
		}
		\caption{MAP estimates of the distribution parameters} \label{alg:MAP}
	\end{algorithm}

	\subsection{Probability distribution of rates} \label{sec:rate_dist}
	As we defined in Section \ref{sec:rate}, the execution rate function belongs to a class of real-valued functions,
	\begin{equation*}
	\Gamma = \{ r: [0, 1] \mapsto \mathbb{R}\}.
	\end{equation*}
	We define a probability space of the real functions using a Gaussian process. Suppose that the execution rate data, $\{r_{m}(t_l); m = 1,\ldots, M, l = 1,\ldots, L\}$, are the noisy observations of an unknown rate function $r \in \Gamma$,
	\begin{equation*}
	r_m(t_l) = r(t_l) + \epsilon_{ml},
	\end{equation*}
	where $\epsilon_{ml} \sim \mathcal{N}(0, \sigma^{r})$. We assume the unknown rate function $r$ is a realization of Gaussian process with zero mean and covariance function $\mathcal{Q}(\cdot, \cdot)$. We would like to estimate the posterior distribution of an execution rate function evaluated at a test time point $t$, which is denoted by $r(t)$. Let $\V{r}$ denote a $ML \times 1$ vector with $r_m(t_l)$ as its $m+(l-1)M$th element. By the definition of the Gaussian process, the joint distribution of the data $\V{r}$ and $r(t)$ is a multivariate normal distribution,
	\begin{equation*}
	\left[ \begin{array}{c} \V{r} \\ r(t)
	\end{array} \right] \sim \mathcal{N}\left(\left[ \begin{array}{c} \V{0} \\ 0
	\end{array} \right], \left[\begin{array}{c c}
	\sigma^2_{r}\M{I} + \M{Q}_{\V{r}, \V{r}} & \V{q}_{\V{r}, r} \\
	\V{q}_{\V{r}, r}^T  & q_{r, r} 
	\end{array} \right] \right),
	\end{equation*}
	where $\M{Q}_{\V{r}, \V{r}}$ is a $ML \times ML$ matrix with $\mathcal{Q}(r_{m1}(t_{l1}), r_{m2}(t_{l2}))$ as its $(m1+(l1-1)M, m2+(l2-1)M)$th element, $\M{q}_{\V{r}, r}$ is a $ML \times 1$ vector with $\mathcal{Q}(r_{m1}(t_{l1}), r(t))$ as its $m1+(l1-1)M$th element, and $q_{r, r}  = \mathcal{Q}(r(t), r(t))$. By the Gaussian conditioning formula, the posterior distribution of $r(t)$ is a normal distribution with the following mean and variance,
	\begin{equation}
	\begin{split}
	& E[r(t)|\V{r}] = \V{q}_{\V{r}, r}^T (\sigma^2_{r}\M{I} + \M{Q}_{\V{r}, \V{r}})^{-1} \V{r} \\
	& Cov[r(t)|\V{r}] = q_{r, r}  - \V{q}_{\V{r}, r}^T (\sigma^2_{r}\M{I} + \M{Q}_{\V{r}, \V{r}})^{-1} \V{q}_{\V{r}, r}.
	\end{split}
	\end{equation}
	
	\subsection{Sufficient dimension reduction of postures and posture sequences} \label{sec:edr}
	We are interested in investigating how the variation of $\tilde{\alpha}(t)$ is related to the rate $r(t)$ and what direction of the variation is strongly correlated to the rate. For doing so, we borrow the concept of the effective dimension reduction \citep{li1991sliced}. 
	
	Let $\V{y}_t$ represent a vectorization of the posture $\tilde{\alpha}(t)$, and let $\V{c}_t \in \mathbb{R}^{2(n-1)}$ denote the corresponding tangent coordinate sampled from the density \eqref{eq:xlike}. The corresponding rate is denoted by $r_t = r(t)$. Assume that $r_t$ is an unknown regression function of $\V{c}_t$,
	\begin{equation*}
	r_t = f(\V{\beta}^T_1 \V{c}_t, \ldots, \V{\beta}^T_B \V{c}_t, \epsilon),
	\end{equation*}
	where $\V{\beta}_b \in \mathbb{R}^{2(n-1)}$ is an unknown projection vector for $b = 1,\ldots, B$, and $\epsilon$ is independent of $\V{c}_t$. The $B$ projection features, $\V{\beta}^T_1 \V{c}_t, \ldots, \V{\beta}^T_B \V{c}_t$, are sufficient statistics to estimate $f$, because $\V{c}_t$ and $r_t$ are conditionally independent given the projection features. 
	
	We are interested in identifying the $B$ projection vectors with $M$ observed pairs of $\V{c}_t$ and $r_t$, which are denoted by $\{\V{c}_{m,t}, r_{m,t}; m = 1,\ldots, M\}$. Using Theorem 3.1 of \citet{li1991sliced} with the elliptical symmetry of the density \eqref{eq:xlike} and $E[\V{c}_t]= \V{0}$, it is easy to show
	\begin{equation} \label{eq:sir}
	E[\V{c}_t|r_t] = \sum_{b=1}^B \phi_b \V{K}_t \V{\beta}_b,
	\end{equation}
	where $\M{K}_t = cov[\V{c}_t]$, and $\phi_b \in \mathbb{R}$ is an unknown coefficient. Therefore, one can identify $\{\M{K}_t\V{\beta}_b; b=1,\ldots, B\}$ by taking the $B$ largest eigenvectors of the covariance matrix $cov[E[\V{c}_t|r_t]]$, and $\V{\beta}_b$ can be achieved by multiplying $\M{K}_t\V{\beta}_b$ with $\M{K}_t^{-1}$. Estimating the covariance matrix $cov[E[\V{c}_t|r_t]]$ is not difficult. The expectation $E[\V{c}_t|r_t]$ can be estimated at discrete points, $\{r_{mt}; m=1,\ldots,M\}$, by using the local kernel regression,
	\begin{equation*}
	\hat{E}[\V{c}_t|r_t] = \frac{\sum_{m=1}^M \V{c}_{m,t} K_h(|r_t- r_{mt}|) }{\sum_{m=1}^M K_h(|r_t- r_{mt}|) }.
	\end{equation*}
	The estimate of the covariance matrix $cov[E[\V{c}_t|r_t]]$ can be taken as the sample covariance of the estimated expectations, 
	\begin{equation} \label{eq:sir2}
	\hat{cov}[E[\V{c}_t|r_t]] = \frac{1}{M} \sum_{m=1}^M  \hat{E}[\V{c}_t|r_{mt}] (\hat{E}[\V{c}_t|r_{mt}])^T. 
	\end{equation}
	Estimating the unconditional covariance $\V{K}_t$ can be achieved by Algorithm \ref{alg:MAP}. 
	
	The same effective dimension reduction can be easily extended for identifying the variation direction of posture sequences strongly related to the rate of motion. Suppose that we have observations of temporally aligned posture sequences at $L$ discrete time points, $\{t_1, t_2, \ldots, t_L\}$. Let $\V{c}_{s:t} = (\V{c}_{t_1}^T, \V{c}_{t_2}^T, \ldots, \V{c}_{t_L}^T)^T$ denote a long vector concatenating the tangent coordinates of the postures observed from time $s$ to time $t$. The corresponding rate in between time $s$ and time $t$ can be defined as
	\begin{equation*}
	r_{s:t} = \log\left(\int_{s}^t \exp\{r(u)\} du \right),
	\end{equation*}
	which is the logarithm of the total physical time spent to perform the posture $\tilde{\alpha}(s)$ to $\tilde{\alpha}(t)$ in the given sequence. The effective dimension reduction can be defined by replacing $\V{c}_t$ and $r_t$ with $\V{c}_{s:t}$ and $r_{s:t}$ in the the effective dimension reduction for postures. \\

	
	\section{Applications to operations research} \label{sec:app}
	We demonstrate five use cases of the proposed methodology for the motion and time analysis in manufacturing operations. Due to the page limitation, this section presents only three of the use cases, and the remaining two use cases will be presented in e-companion \ref{EC3}. The first subsection describes the dataset that we use for the use cases, and the other three subsections demonstrate the three use cases, one per section.
	
	\subsection{Dataset: Manufacturing Operational Motions} \label{sec:dataset}
\begin{figure}[t]
		\centering
		\includegraphics[width=\textwidth]{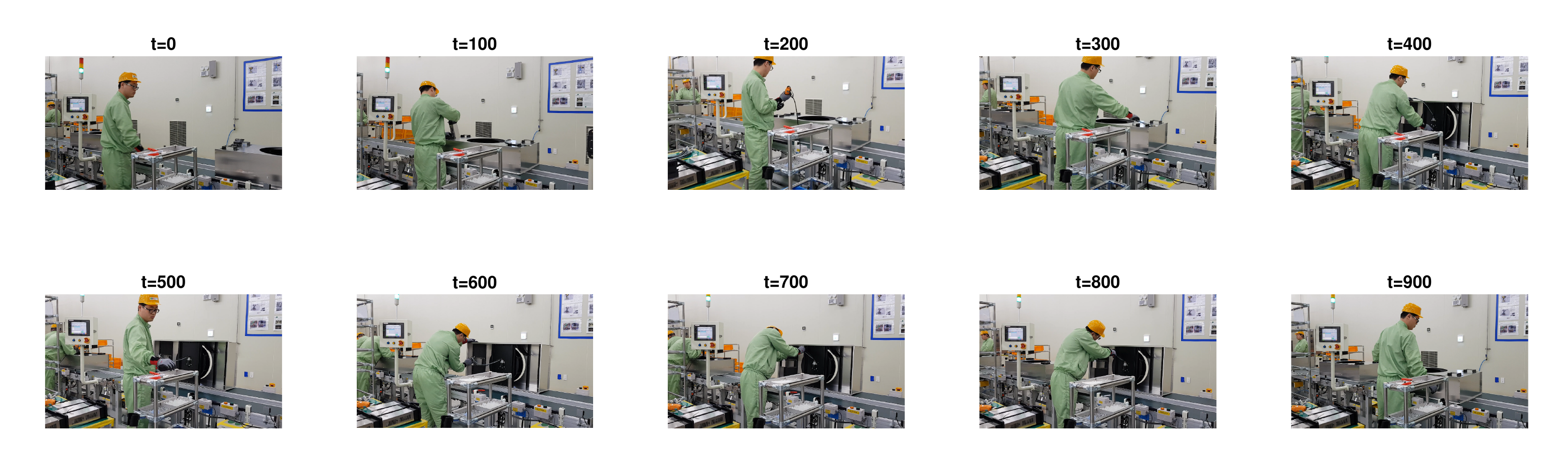}
		\caption{Snapshot images of the manufacturing operation that we study}
		\label{fig8}
	\end{figure}

	We applied the proposed statistical analysis to analyze manufacturing operations in a factory. Five different operations are measured and analyzed. Figure \ref{fig8} shows some snapshot images of the first operation, where a worker stands in front of manufacturing workbenches connected through a convey belt line. This illustrates a typical example of manufacturing operations. A human operator stands with not much change in the standing location, mostly using the upper body to pick tools and do works. 
	
	For each operation, a standard operating procedure is first produced in the form of a video recording of an experienced worker's performance, and other workers are trained with the work manual. Each of the five operations is performed 60 times by different workers, while the motions are recorded using the Perception Neuron Motion Capture\texttrademark. The motion sensor periodically measures the skeleton data with 21 landmark coordinates as described in Figure \ref{fig1} with a constant measurement interval, which produces a sequence of skeleton data. In total, we have $60$ sequences of skeleton data for each of the five operation types, including one corresponded to the standard operating procedure. In total, 300 motion data are analyzed.
	
	\subsection{Use Case 1: Motion Recognition} \label{sec:use1}
	Motion recognition is a problem of inferring what humans perform from an observed motion sequence. The most existing works in computer vision have been focused on the motion recognition. In the context of operations research, it is useful to automatically identify the operation types currently being performed. In this section, we demonstrate how our foundational work in the posture space and motion space is applied to the human motion recognition. The results are compared to those from a deep learning approach. 
	
	The raw data for the 300 motion sequences are available in the form of skeleton sequences. We first converted the skeleton data to posture sequences, following our definition of human postures in Section \ref{sec:posture}. Let $\alpha_m \in \mathcal{A}$ represent the posture sequence of the $m$th skeleton data for $m=1,\ldots,300$. For each pair of the posture sequences, $\alpha_{m1}$ and $\alpha_{m2} \in \mathcal{A}$, we can compute the motion distance, $d_{\mathcal{H} / \mathcal{G}}([\alpha_{m1}], [\alpha_{m2}])$. The statistics of the distances between two motions from the same operation type and two motions from different operations types are summarized in Table \ref{tbl1}. The distances between different operations types are significantly larger than those from the same operation types. 
	
	\begin{table}
		\centering
		\begin{tabular}{|c|c c c c c|}
			\hline
			Operation & Op1 & Op2 & Op3 & Op4 & Op5\\
			\hline
			Op1& 5.62 ($\pm$2.00)& 12.62 ($\pm$1.37)& 8.23 ($\pm$0.74)& 10.83 ($\pm$1.82)& 10.40 ($\pm$1.89) \\
			Op2& 12.62 ($\pm$1.37)& 6.23 ($\pm$2.97)& 12.08 ($\pm$1.35)& 10.21 ($\pm$1.20)& 10.14 ($\pm$1.78)\\
			Op3& 8.23 ($\pm$0.74)& 12.08 ($\pm$1.35)& 5.09 ($\pm$1.82)& 9.78 ($\pm$2.06)& 9.87 ($\pm$2.07)\\
			Op4& 10.83 ($\pm$1.82)& 10.21 ($\pm$1.20)& 9.78 ($\pm$2.06)& 7.13 ($\pm$3.99)& 8.75 ($\pm$1.98)\\
			Op5& 10.40 ($\pm$1.89)& 10.14 ($\pm$1.78)& 9.87 ($\pm$2.07)& 8.75 ($\pm$1.98)& 6.69 ($\pm$3.56)\\
			\hline
		\end{tabular}
		\caption{Motion distances $d_{\mathcal{H} / \mathcal{G}}([\alpha_{m1}], [\alpha_{m2}])$ between different operation types of Op1, Op2, Op3, Op4 and Op5. The number in each cell is the average distance, and the number in the round bracket is the standard deviation.}
		\label{tbl1}
	\end{table}
	
	Based on this distinction, we can build a simple nearest neighborhood classifier to infer the operation type of a posture sequence. We randomly selected 80\% of the 300 posture sequences as training data, and the remaining 20\% is reserved as testing data. The training data come with 240 postures sequences labeled with the associated operations types. For each posture sequence in the test data, the nearest neighborhood classifier assigns its operation type to the operation type of the closest posture sequence in the training data, where the `closedness' is defined in terms of the motion distance $d_{\mathcal{H}/\mathcal{G}}$. The assigned operation type is compared to the true operation type of the test posture sequence for evaluating the accuracy of the classifier. For all test cases, the operation types are correctly identified with 100\% accuracy. As a benchmark, we fit the Deep Learning Long Short-Term Memory (LSTM) algorithm to the skeleton data with the same split of training and test data, and its classification accuracy is 95\%. This shows that our motion definition $[\alpha]$ and the corresponding distance $d_{\mathcal{H} / \mathcal{G}}$ are effective in distinguishing different human motions. 
	
	\subsection{Use Case 2: Analysis of Rate Variations and Bottleneck Analysis} \label{sec:use2}
\begin{figure}[t]
		\centering
		\includegraphics[width=\textwidth]{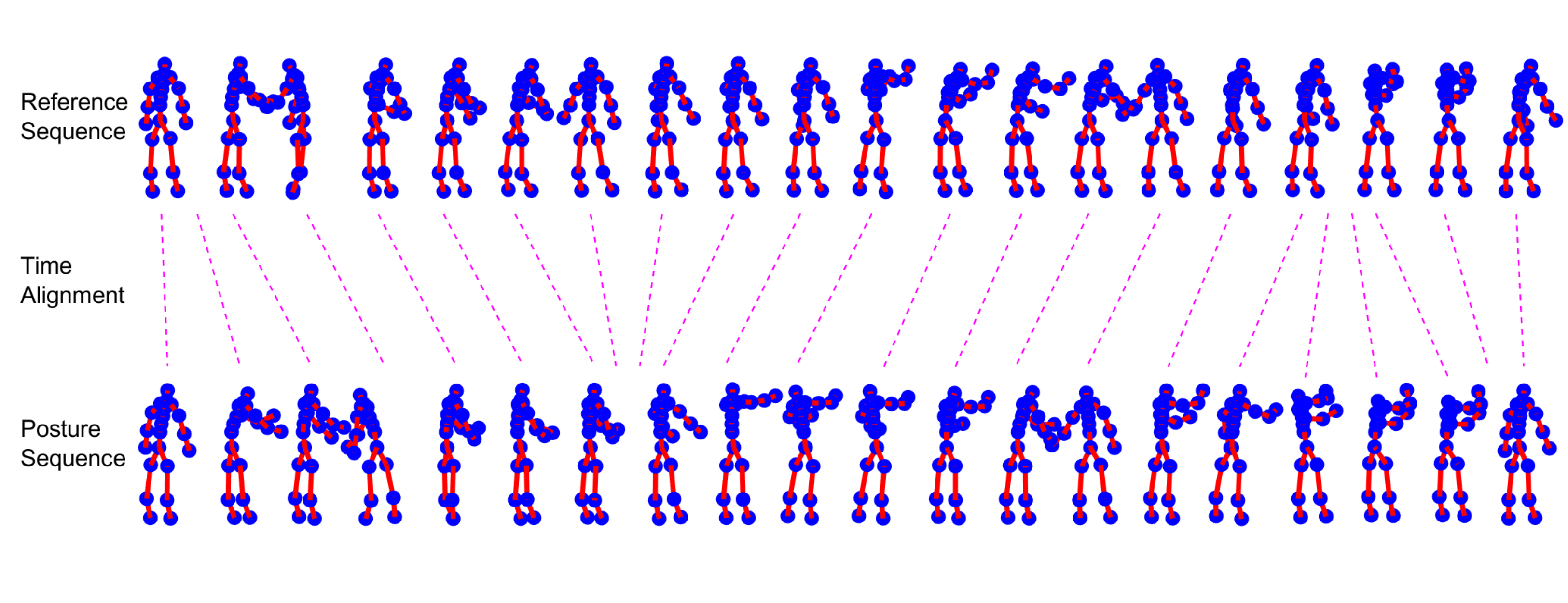}
		\caption{Time Alignment of a Posture Sequence $\alpha_m$ to a Reference Posture Sequence. The top panel shows the reference sequence, and the bottom panel shows the posture sequence $\alpha_m$. In the middle, the dotted lines shows how the time-alignment function $\gamma_m$ maps from the reference sequence to $\alpha_m$.}
		\label{fig6}
	\end{figure}

\begin{figure}[ht!]
		\includegraphics[width=\textwidth]{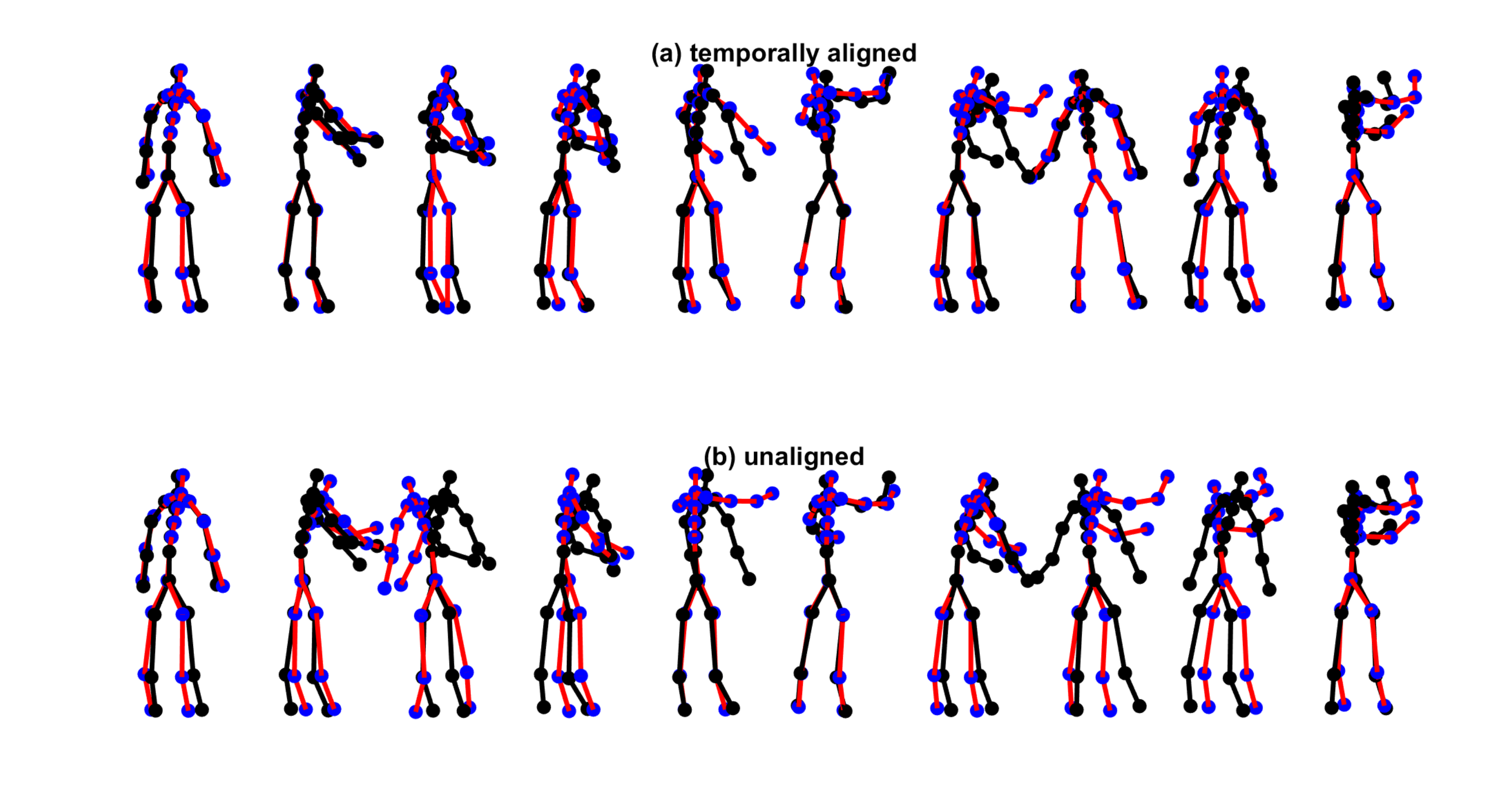}
		\caption{Effectiveness of the Time Alignment of a Posture Sequence to a Reference Posture Sequence. Panel (a) shows the time-aligned posture sequence (red and blue) to the reference posture sequence (black), and panel (b) shows the unaligned posture sequence. }
		\label{fig3}
	\end{figure}

	In this subsection and all subsequent subsections, we chose one of the five operation types to demonstrate advanced motion analysis. We chose the first operation type. We present the analysis of variations in execution rates and a bottleneck analysis that identifies critical work elements that slow down the progression of the operation most. To define the relative work execution rates for the 60 motions of the first operation type, we set the posture sequence representing the standard operating procedure (SOP) as a reference posture sequence $\alpha_R \in \mathcal{A}$, and the other posture sequences are temporally aligned to the reference sequence as described in Section \ref{sec:stat},
	\begin{equation*}
	\gamma_m = \argmin_{\gamma \in \mathcal{G}} d_{\mathcal{H}}(h_{\alpha_m \circ \gamma_m}, h_{\alpha_R}).
	\end{equation*}
	This time alignment step basically finds the time-alignment function $\gamma_m$ that temporally aligns $\alpha_m$ to $\alpha_R$. Figure \ref{fig6} illustrates the time-alignment function for a selected posture sequence (shown in the bottom panel) to the reference posture sequence (shown in the top panel). Figure \ref{fig3} illustrates the effectiveness of the time alignment by overlaying the aligned sequence on the reference sequence. The top panel of the figure matches the temporally aligned posture sequence $\alpha_m \circ \gamma_m$ versus the reference sequence $\alpha_R$, and the bottom panel shows $\alpha_m$ versus $\alpha_R$. The temporal alignment is very accurate. 
	
	\begin{figure}[ht]
		\centering
		\includegraphics[width=0.9\textwidth]{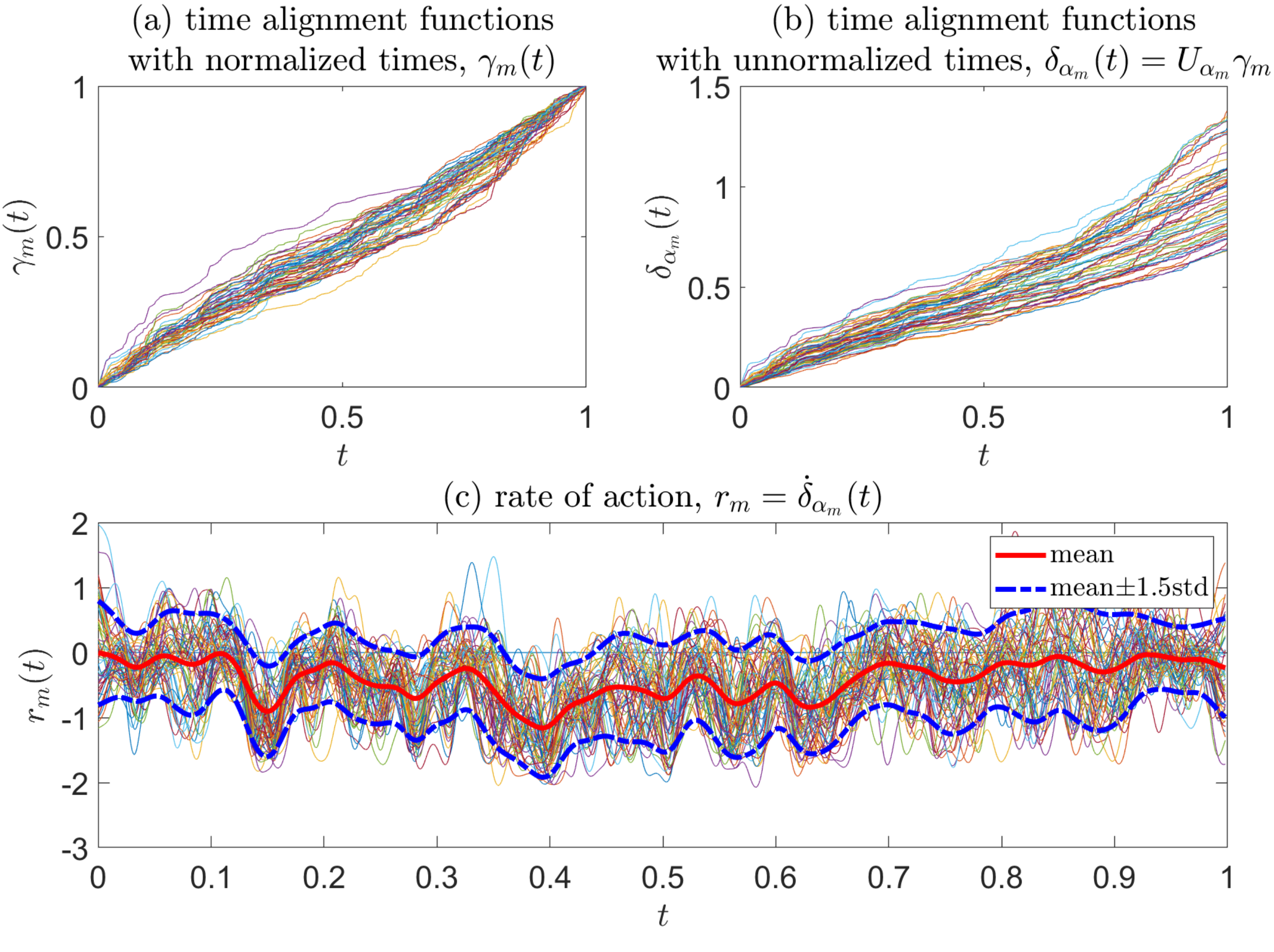}
		\caption{Time alignment functions for 60 performances of the first operation type. }
		\label{fig4}
	\end{figure}
	As discussed in Section \ref{sec:rate}, the time-alignment function $\gamma_m$ captures the execution rate of posture sequence $\alpha_m$ relative to the reference posture sequence. Analyzing the time-alignment functions over the 60 posture sequences would identify the temporal variability of the first operation type, and we can identify the work elements that run at slower paces, which can be interpreted as bottleneck work elements. Please note that the time-alignment functions are defined at the normalized time scale. They are first scaled to the time-alignment functions defined at the physical times to $\delta_{\alpha_m} = U_{\alpha_m}\gamma_m$. We took the derivative of the scaled function to get the work rate function, $r_m(t) = \log(\dot{\delta}_{\alpha_m}(t))$. Figure \ref{fig4}-(a) shows the time alignment functions at the normalized time scale, and Figure \ref{fig4}-(b) shows the corresponding functions at the physical time scale, while Figure \ref{fig4}-(c) shows the work rate functions. The rate function for the reference posture sequence is a zero function. A positive value of $r_m(t)$ implies that the instant speed of the $m$th work sequence at time $t$ is faster than that of the reference posture sequence. Conversely, if $r_m(t) < 0$, the instant speed is slower than that of the reference posture sequence. The bottleneck is identified by analyzing the rate function values within a time window centered at each test time location and identifying $t^*$ satisfying 
	\begin{equation} \label{eq:bottleneck}
	t^* = \argmin_{t \in \{t_1, \ldots, t_L\}} \sum_{|t_l-t|<\Delta} \sum_{m=1}^M \min\{0, \gamma_m(t_l)\},
	\end{equation}
	where $\Delta$ is a positive constant that represents the size of a time window. When $\Delta = 0.020$, the identified bottleneck corresponds to the work elements performed at time $t^* = 0.384$ in the normalized time scale. The work elements performed in the time window centered at $t^*$ will be further analyzed to identify the best practice in Section \ref{sec:use4}.
	
	\subsection{Use Case 3: Best Practice Analysis} \label{sec:use5}
\begin{figure}[ht!]
		\centering
		\includegraphics[width=\textwidth]{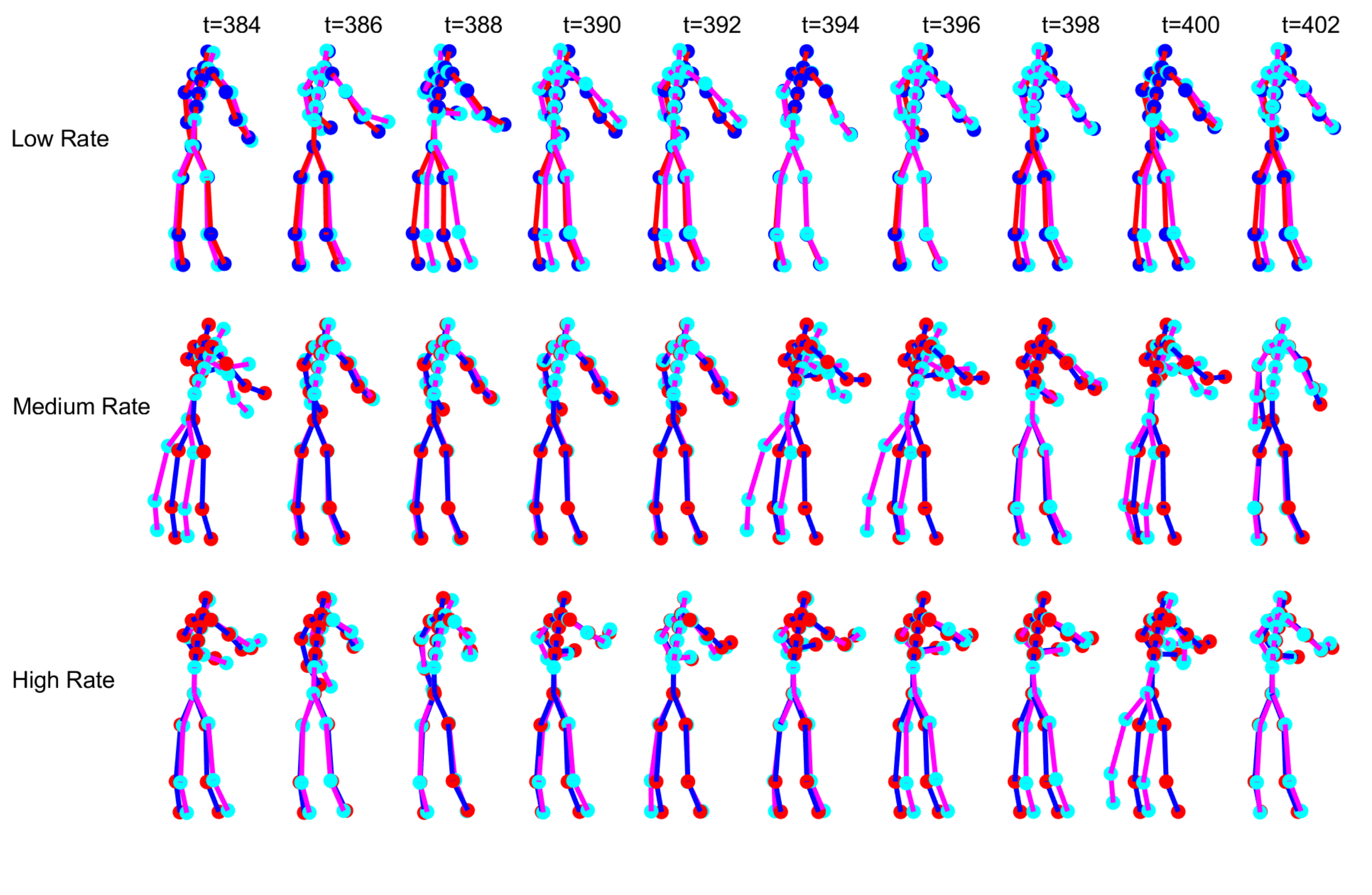}
		\caption{Estimated work sequences versus observed work sequences for high, medium and low work rates. The estimated ones are depicted as blue dots connected through red lines, and the observed ones are with cyan dots connected through magenta lines. }
		\label{fig5}
	\end{figure}

	We analyzed the subsequences of the 60 rate-normalized posture sequences from the first operation type around the bottleneck process time $t^*$ identified by \eqref{eq:bottleneck},  
	\begin{equation*}
	\tilde{A} = \{ \tilde{\alpha}_m(t_l); m = 1,\ldots, M, t \in |t_l-t^*|<\Delta\},
	\end{equation*} 
	and the corresponding rate function values
	\begin{equation*}
	\tilde{R} = \{ r_m(t_l); m = 1,\ldots, M, t \in |t_l-t^*|<\Delta\}.
	\end{equation*} 
	We applied the effective dimension reduction described in Section \ref{sec:edr} and identified the projection directions of the subsequences that are correlated to the rate function values most. Two projection directions are identified, which explain more than 90\% of the variation of the subsequences. A simple linear regression model is fit to relate the two projection features to the rate functions values. Based on the regression model, we can identify the values of the two projections features that give low, medium and high rate function values. We reconstruct the corresponding work subsequences using the projection features. Figure \ref{fig5} shows the reconstructed subsequences for low, medium and high work rates, compared to the observed work subsequences giving the same rate function values. The reconstructed ones match well to the corresponding observations. This means that the sufficient dimension reduction and the regression analysis is effective. Comparing the work sequences for different work rates, we can see there are significant differences in hand poses. In the sequence with a high work rate, two hand positions are around the chest level, and two arms are bended so as to have the main body closer to the two hands.  
	
	\section{Conclusion} \label{sec:conc}
	We have discussed an important data science problem of analyze modern motion sensor data for an automated motion and time study as to analysis of variations in work motions and work execution rates. Many existing studies have been focused only on motion analysis. This paper is the first of the kind in analyzing work motions and rates together for a work evaluation and performance analysis. We contributed many mathematical developments relevant. We defined a mathematical notion of a human posture, based on the skeletonized description of a human body, which considers the unwanted variations of the human skeleton data due to person-to-person variations in body parts and absolute locations. We defined a Riemannian manifold to build the space of human postures and defined a probability distribution over the manifold space of human postures. We also defined the notion of a motion as a sequence of human postures sorted in the orders of their executions, and the metric space and probability distribution over the metric space were built on the base of the space and probability distribution of human postures. Finally, we defined the relative execution rate of a motion as a function that temporally aligns the given motion to a reference motion, and the space and probability distribution of the rate functions are defined. Correlation of the variations of motions and corresponding execution rates are sought. Those foundational developments are highlighted with five use cases applied to manufacturing motion data, including automatic work recognition, bottleneck work identification, analysis of motion variations, work standardization and best practice analysis. 
	
	We expect this research impactful on many practical areas. This work is initially motivated by quantifying the variations and uncertainties of human works performed as one of many sequential production steps in a factory. In the modern manufacturing, many of the production steps are automated with robotic operations, but still quite a large number of steps are being performed by humans. Understanding the variation of human works and the impact of the variation to the subsequent steps is important to quantify and improve the performance of the production line. Our mathematical and statistical developments along with Use Case 4 would provide the uncertainty quantification of human works. Bottleneck analysis and subsequent best practice analysis are important steps to the motion and time study, but they have been performed ineffectively with potential human biases. Our methodological developments with modern motion sensing would automate the inefficient process, producing quantitative outcomes. Moreover, the best practice analysis outcome will be potentially exploit to train industrial robotics with the best human operations identified. Our methodological developments are general enough, so we believe they will find other use cases that we have not discussed.
	
	\ACKNOWLEDGMENT{We acknowledge support for this work from the Brain Pool Fellowship of the National Research Foundation Korea (grant no. 2019H1D3A2A01100649).}

	\bibliographystyle{informs2014} 
	\bibliography{MOTION} 
	
	\ECSwitch
	\ECHead{Derivation of Statistical Inference Algorithms and Extra Use Cases}
	The e-companion largely consists of two parts. Sections \ref{EC1} and \ref{EC2} contain the detailed derivation of the statistical inference algorithms that presented in Algorithm \ref{alg:MLE} and Algorithm \ref{alg:MAP}. Section \ref{EC3} presents two extra use cases of the proposed methodology. 
	
	\section{Derivation of ML Estimation Algorithm \ref{alg:MLE}} \label{EC1}
	Given the relevant data, $\mathcal{D}_l = \{\V{y}^{(m,l)}; m = 1,\ldots, M\}$, the negative log likelihood function of $\V{\mu}^{(l)}$ and $\V{K}_l$ is
	\begin{equation*}
	\begin{split}
	\mathcal{L}(\V{\mu}^{(l)}, \V{K}_l) =& \sum_{m=1}^M -2\log p(\V{c}^{(m,l)}|\V{\mu}^{(l)}, \M{K}_l) \\
	=&  M \log|\M{K}_l| + \sum_{m=1}^M  (\V{c}^{(m,l)})^T \M{K}_l^{-1} \V{c}^{(m,l)}.
	\end{split}
	\end{equation*}
	We first present an iterative algorithm to achieve the maximum likelihood estimate (MLE) of $\V{\mu}^{(l)}$, because the MLE of $\M{K}_l$ is relatively easy to achieve if the MLE of $\V{\mu}^{(l)}$ is given. Let $\V{\mu}^{(l,j)}$ represent the solution achieved at iteration $j$ of the iterative algorithm. We can take the tangent space of $\mathcal{Y}$ at $\V{\mu}^{(l,j)}$, and let $\V{c}^{(m,l,j)}$ denote the tangent coordinate of $\V{y}^{(m,l)}$ in the tangent space, which is equal to
	$$  \V{c}^{(m,l,j)}= \M{W}_{\V{\mu}^{(l,j)}}^T \V{y}^{(m,l)}.$$

	 At iteration $j+1$, the estimate of $\V{\mu}^{(l)}$ is updated with $\V{\mu}^{(l,j+1)}$. Accordingly, we can take the tangent space of $\mathcal{Y}$ at the updated mean $\V{\mu}^{(l,j+1)}$ and take the tangent coordinate of $\V{y}^{(m,l)}$ in the tangent space, denoted by $\V{c}^{(m,l,j+1)}$. To relate $\V{c}^{(m,l,j)}$ and $\V{c}^{(m,l,j+1)}$, we use the first order Taylor expansion, 
	\begin{equation} \label{eq:firstorder}
	\begin{split}
	\V{c}^{(m,l,j+1)} &\approx \M{W}_{\V{\mu}^{(l,j)}}^T(\V{y}^{(m,l)} - \V{\mu}^{(l,j+1)}) \\
	&= \V{c}^{(m,l,j)} - \M{W}_{\V{\mu}^{(l,j)}}^T\V{\mu}^{(l,j+1)}.
	\end{split}
	\end{equation}
	It is easy to check that as $\V{\mu}^{(l,j)} \rightarrow \V{\mu}^{(l,j+1)}$, the rhs of \eqref{eq:firstorder} goes to 
	\begin{equation*}
	\begin{split}
	\lim_{\V{\mu}^{(l,j)} \rightarrow \V{\mu}^{(l,j+1)}} \M{W}_{\V{\mu}^{(l,j)}}^T(\V{y}^{(m,l)} - \V{\mu}^{(l,j+1)}) &= \M{W}_{\V{\mu}^{(l,j+1)}}^T(\V{y}^{(m,l)} - \V{\mu}^{(l,j+1)}) \\
	&= \M{W}_{\V{\mu}^{(l,j+1)}}^T\V{y}^{(m,l)} - \M{W}_{\V{\mu}^{(l,j+1)}}^T\V{\mu}^{(l,j+1)} \\
	& = \V{c}^{(m,l,j+1)}. 
	\end{split}
	\end{equation*}
	
	At iteration $j+1$, we choose $\V{\mu}^{(l,j+1)}$ that minimizes the negative log likelihood term,
	\begin{equation*}
	\begin{split}
	\V{\mu}^{(l,j+1)} &= \arg\min \left\{ \sum_{m=1}^M  (\V{c}^{(m,l,j+1)})^T \M{K}_l^{-1} \V{c}^{(m,l,j+1)} \right\}\\
	                  &= \arg\min \left\{ \sum_{m=1}^M  (\V{c}^{(m,l,j)} - \M{W}_{\V{\mu}^{(l,j)}}^T\V{\mu})^T \M{K}_l^{-1} (\V{c}^{(m,l,j)} - \M{W}_{\V{\mu}^{(l,j)}}^T\V{\mu})\right\}
	\end{split}
	\end{equation*}
	The first order necessary condition for optimality is
	\begin{equation} \label{eq:mu_mle}
	\begin{split}
	\M{W}_{\V{\mu}^{(l,j)}}^T\V{\mu}^{(l,j+1)} = \frac{1}{M} \sum_{m=1}^M \V{c}^{(m,l,j)},
	\end{split}
	\end{equation}
	and $\V{\mu}^{(l,j+1)}$ can be achieved from the condition,
	\begin{equation}
	\V{\mu}^{(l, j+1)} = \exp_{\V{\mu}^{(l,j)}}(\tilde{\M{W}}_{\V{\mu}^{(l,j)}}\M{W}_{\V{\mu}^{(l,t)}}^T \V{\mu}^{(l, j+1)}),
	\end{equation}
	where $\tilde{\M{W}}_{\V{\mu}^{(l,j)}}$ is a block diagonal matrix with the $i$th diagonal as $\M{W}_{\V{\mu}^{(l,j)}_i}$. This iteration updates the mean estimate $\V{\mu}^{(l,j)}$ iteratively, which continues until the convergence, which can be checked with the magnitude of $||\M{W}_{\V{\mu}^{(l,j)}}^T\V{\mu}^{(l,j+1)}||_F$. When $\V{\mu}^{(l,j+1)} = \V{\mu}^{(l,j)}$, the norm is zero. As the two successive means get very close, the norm decreases. We can check whether the norm is less than a chosen threshold $\xi_1$ to define the convergence. The solution at the convergence is the maximum likelihood estimate $\hat{\V{\mu}}^{(l)}$. 
	
	Suppose that the convergence is achieved at iteration $j^*$. The maximum likelihood estimate for the covariance parameter can be achieved by
	\begin{equation} \label{eq:K_mle}
	\hat{\M{K}}_l = \frac{1}{M} \sum_{m=1}^M \V{c}^{(l,m,j^*)} (\V{c}^{(l,m,j^*)})^T.
	\end{equation}
	
	\section{Derivation of MAP Estimation Algorithm \ref{alg:MAP}} \label{EC2}
	We assume that the means are temporally correlated through the following autocorrelation model: for $l = 1,\ldots, L$,  
	\begin{equation*}
	\begin{split}
	p(\V{\mu}^{(l)}|\V{\mu}^{(l-1)}, \lambda_0^2 \M{I}) =& a_0 \lambda_0^{1-n} \exp\left\{ - \frac{1}{2\lambda_0^2}(\V{\mu}^{(l)})^T \M{W}_{\V{\mu}^{(l-1)}} \M{W}_{\V{\mu}^{(l-1)}}^T \V{\mu}^{(l)}  \right\} \\
	& \qquad \prod_{i=1}^{n-1} \left[\frac{\theta_{\V{\mu}^{(l-1)}_i}}{\sin(\theta_{\V{\mu}^{(l-1)}_i})}  I_{\left|\left|\theta_{\V{\mu}^{(l-1)}_i}\right|\right|\le \pi/2} \right],
	\end{split}
	\end{equation*}
	where $\lambda_0^2$ is the overall variance parameter, and $\V{\mu}^{(0)} \in \mathcal{Y}$ is the mean of $\V{\mu}^{(1)}$. Noting that $(\V{\mu}^{(l)})^T \M{W}_{\V{\mu}^{(l-1)}} \M{W}_{\V{\mu}^{(l-1)}}^T \V{\mu}^{(l)} = \sum_{i=1}^{n-1} d_{\mathcal{S}^2}(\V{\mu}^{(l)}_i, \V{\mu}^{(l-1)}_i)^2$ and $\theta_{\V{\mu}^{(l-1)}_i} = d_{\mathcal{S}^2}(\V{\mu}^{(l)}_i, \V{\mu}^{(l-1)}_i)$, the density function can be written as
	\begin{equation*}
	\begin{split}
	p(\V{\mu}^{(l)}|\V{\mu}^{(l-1)}, \lambda_0^2 \M{I}) =& a_0 \lambda_0^{1-n} \exp\left\{ - \frac{1}{2\lambda_0^2}\sum_{i=1}^{n-1} (d_i^{(l-1, l)})^2 \right\} \\
	& \qquad \prod_{i=1}^{n-1} \left[\frac{d_i^{(l-1, l)}}{\sin(d_i^{(l-1, l)})}  I_{\left|\left|d_i^{(l-1, l)}\right|\right|\le \pi/2} \right],
	\end{split}
	\end{equation*}
	where $d_i^{(l-1, l)} = d_{\mathcal{S}^2}(\V{\mu}^{(l)}_i, \V{\mu}^{(l-1)}_i)$. Since $d_i^{(l-1,l)} = d_i^{(l,l-1)}$, 
	\begin{equation*}
	p(\V{\mu}^{(l)}|\V{\mu}^{(l-1)}, \lambda_0^2 \M{I})  = p(\V{\mu}^{(l-1)}|\V{\mu}^{(l)}, \lambda_0^2 \M{I}).
	\end{equation*}
	Moreover, we put the inverse Wishart prior with $\nu_0$ degrees of freedom on each $\M{K}_l$ with its density,
	\begin{equation*}
	p(\M{K}_l|\M{K}_0, \nu_0) = \frac{1}{2^{2(n-1)\nu_0} |\M{K}_0|^{\nu_0/2} \Gamma_{2(n-1)}(\nu_0/2)} |\M{K}_l|^{-(\nu_0+2n-1)/2} \exp\left\{ - \frac{1}{2} tr(\M{K}_0 \M{K}_l^{-1}) \right\}.
	\end{equation*}
	where $\M{K}_0$ is a $2(n-1)$ positive definite matrix. The joint posterior density function is 
	\begin{equation*}
	\begin{split}
	p(\{\V{\mu}^{(l)}, \M{K}_l, l = 1,..,L \} | \mathcal{D}, \lambda_0^2, \V{\mu}_0, \M{K}_0) & \propto \prod_{m=1}^M \prod_{l=1}^L  p(\V{y}^{(m,l)}|\V{\mu}^{(l)}, \M{K}_l) \\
	& \qquad \times \prod_{l=1}^L  p(\V{\mu}^{(l)}|\V{\mu}^{(l-1)}, \lambda_0^2 \M{I}) \\
	& \qquad \times \prod_{l=1}^L  p(\M{K}_l|\M{K}_0).
	\end{split}
	\end{equation*}
	We maximize the joint posterior with respect to the parameter set $\{\V{\mu}^{(l)}, \M{K}_l, l = 1,..,L \}$ to achieve the MAP estimate of the parameter set. As a solution approach, we use the coordinate descent optimization that minimizes the negative log posterior,
	\begin{equation} \label{eq:nlp}
	\begin{split}
	\log p(\{\V{\mu}^{(l)}, \M{K}_l, l = 1,..,L \} | \mathcal{D}, \lambda_0^2, \V{\mu}_0, \M{K}_0) & = -\sum_{m=1}^M \sum_{l=1}^L  \log p(\V{c}^{(m,l)}|\V{\mu}^{(l)}, \M{K}_l) \\
	& \qquad - \sum_{l=1}^L  \log p(\V{\mu}^{(l)}|\V{\mu}^{(l-1)}, \lambda_0^2 \M{I}) \\
	& \qquad - \sum_{l=1}^L  \log p(\M{K}_l|\M{K}_0) \\
	& \qquad + (constant).
	\end{split}
	\end{equation}
	For each $l = 1,...,L$, we optimize $\V{\mu}^{(l)}$ while fixing all the other parameters. Among the terms in the negative log posterior \eqref{eq:nlp}, the following terms only depends on $\V{\mu}^{(l)}$,
	\begin{equation*}
	\begin{split}
	f(\V{\mu}^{(l)})  = & -\sum_{m=1}^M \log p(\V{c}^{(m,l)}|\V{\mu}^{(l)}, \M{K}_l)  - \log p(\V{\mu}^{(l)}|\V{\mu}^{(l-1)}, \lambda_0^2 \M{I}) - \log p(\V{\mu}^{(l+1)}|\V{\mu}^{(l)}, \lambda_0^2 \M{I}) \\
	=& -\sum_{m=1}^M \log p(\V{c}^{(m,l)}|\V{\mu}^{(l)}, \M{K}_l) - \log p(\V{\mu}^{(l-1)}|\V{\mu}^{(l)}, \lambda_0^2 \M{I}) - \log p(\V{\mu}^{(l+1)}|\V{\mu}^{(l)}, \lambda_0^2 \M{I}) \\
	=& \sum_{m=1}^M (\V{y}^{(m,l)})^T \M{W}_{\V{\mu}^{(l)}} \M{K}_l^{-1} \M{W}_{\V{\mu}^{(l)}}^T \V{y}^{(m,l)} \\
	& + \frac{1}{\lambda_0^2} (\V{\mu}^{(l-1)})^T \M{W}_{\V{\mu}^{(l)}} \M{W}_{\V{\mu}^{(l)}}^T \V{\mu}^{(l-1)} + \frac{1}{\lambda_0^2} (\V{\mu}^{(l+1)})^T \M{W}_{\V{\mu}^{(l)}} \M{W}_{\V{\mu}^{(l)}}^T \V{\mu}^{(l+1)} + (constant).
	\end{split}
	\end{equation*}
	We will minimize the objective function by an iterative approach. To describe the iteration, please note that $\M{W}_{\V{\mu}^{(l)}}^T \V{\mu}^{(l)} = 0$, and accordingly the objective function can be written as
	\begin{equation}
	\begin{split}
	f(\V{\mu}^{(l)}) =& \sum_{m=1}^M (\V{y}^{(m,l)}-\V{\mu}^{(l)})^T \M{W}_{\V{\mu}^{(l)}} \M{K}_l^{-1} \M{W}_{\V{\mu}^{(l)}}^T (\V{y}^{(m,l)}-\V{\mu}^{(l)}) \\
	& + \frac{1}{\lambda_0^2} (\V{\mu}^{(l-1)}-\V{\mu}^{(l)})^T \M{W}_{\V{\mu}^{(l)}} \M{W}_{\V{\mu}^{(l)}}^T (\V{\mu}^{(l-1)}-\V{\mu}^{(l)}) \\
	& + \frac{1}{\lambda_0^2} (\V{\mu}^{(l+1)}-\V{\mu}^{(l)})^T \M{W}_{\V{\mu}^{(l)}} \M{W}_{\V{\mu}^{(l)}}^T (\V{\mu}^{(l+1)}-\V{\mu}^{(l)}) + (constant).
	\end{split}
	\end{equation}
	Let $\V{\mu}^{(l,j)}$ denote the solution achieved at iteration $j$. The iteration $j+1$ minimizes 
	\begin{equation*}
	\begin{split}
	\min f_{j+1}(\V{\mu}^{(l, j+1)})  &= \sum_{m=1}^M (\V{y}^{(m,l)} - \V{\mu}^{(l, j+1)})^T \M{W}_{\V{\mu}^{(l,j)}} \M{K}_l^{-1} \M{W}_{\V{\mu}^{(l,j)}}^T (\V{y}^{(m,l)} - \V{\mu}^{(l, j+1)})\\
	& + \frac{1}{\lambda_0^2} (\V{\mu}^{(l-1)} - \V{\mu}^{(l, j+1)})^T \M{W}_{\V{\mu}^{(l,j)}} \M{W}_{\V{\mu}^{(l,j)}}^T (\V{\mu}^{(l-1)} - \V{\mu}^{(l, j+1)}) \\
	& + \frac{1}{\lambda_0^2} (\V{\mu}^{(l+1)} - \V{\mu}^{(l, j+1)})^T \M{W}_{\V{\mu}^{(l,j)}} \M{W}_{\V{\mu}^{(l,j)}}^T (\V{\mu}^{(l+1)} - \V{\mu}^{(l, j+1)}).
	\end{split}
	\end{equation*}
	The solution satisfies the first order necessary condition,
	\begin{equation} \label{eq:mu}
	\M{W}_{\V{\mu}^{(l,j)}}^T \V{\mu}^{(l, j+1)} = (M\M{K}_l^{-1} + 2/\lambda_0^2\M{I})^{-1} \left(\sum_{m=1}^M \M{K}_l^{-1} \M{W}_{\V{\mu}^{(l,j)}}^T \V{y}^{(m,l)}  +  \frac{1}{\lambda_0^2} \M{W}_{\V{\mu}^{(l,j)}}^T (\V{\mu}^{(l+1)}+\V{\mu}^{(l-1)}) \right).
	\end{equation}
	Thus, the iteration $j+1$ first achieves $\M{W}_{\V{\mu}^{(l,j)}}^T \V{\mu}^{(l, j+1)}$, and $\V{\mu}^{(l, j+1)}$ is achieved by taking the exponential map, 
	\begin{equation}
	\V{\mu}^{(l, j+1)} = \exp_{\V{\mu}^{(l,j)}}(\tilde{\M{W}}_{\V{\mu}^{(l,j)}}\M{W}_{\V{\mu}^{(l,t)}}^T \V{\mu}^{(l, j+1)}),
	\end{equation}
	where $\tilde{\M{W}}_{\V{\mu}^{(l,j)}}$ is a block diagonal matrix with the $i$th diagonal as $\M{W}_{\V{\mu}^{(l,j)}_i}$. 
	
	At the convergence of the iterations, $\V{\mu}^{(l,j)} = \V{\mu}^{(l,j+1)}$, which satisfies
	\begin{equation*}
	\begin{split}
	\min f_{j+1}(\V{\mu}^{(l, j+1)})  &= \sum_{m=1}^M (\V{y}^{(m,l)} - \V{\mu}^{(l, j+1)})^T \M{W}_{\V{\mu}^{(l,j+1)}} \M{K}_l^{-1} \M{W}_{\V{\mu}^{(l,j+1)}}^T (\V{y}^{(m,l)} - \V{\mu}^{(l, j+1)})\\
	& + \frac{1}{\lambda_0^2} (\V{\mu}^{(l-1)} - \V{\mu}^{(l, j+1)})^T \M{W}_{\V{\mu}^{(l,j+1)}} \M{W}_{\V{\mu}^{(l,j+1)}}^T (\V{\mu}^{(l-1)} - \V{\mu}^{(l, j+1)}) \\
	& + \frac{1}{\lambda_0^2} (\V{\mu}^{(l+1)} - \V{\mu}^{(l, j+1)})^T \M{W}_{\V{\mu}^{(l,j+1)}} \M{W}_{\V{\mu}^{(l,j+1)}}^T (\V{\mu}^{(l+1)} - \V{\mu}^{(l, j+1)}) \\
	&= \sum_{m=1}^M (\V{y}^{(m,l)})^T \M{W}_{\V{\mu}^{(l,j+1)}} \M{K}_l^{-1} \M{W}_{\V{\mu}^{(l,j+1)}}^T \V{y}^{(m,l)}\\
	& + \frac{1}{\lambda_0^2} (\V{\mu}^{(l-1)})^T \M{W}_{\V{\mu}^{(l,j+1)}} \M{W}_{\V{\mu}^{(l,j+1)}}^T \V{\mu}^{(l-1)} \\
	& + \frac{1}{\lambda_0^2} (\V{\mu}^{(l+1)})^T \M{W}_{\V{\mu}^{(l,j+1)}} \M{W}_{\V{\mu}^{(l,j+1)}}^T \V{\mu}^{(l+1)}.
	\end{split}
	\end{equation*}
	Therefore, $\V{\mu}^{(l,j+1)}$ should be the minimizer of $f(\V{\mu}^{(l)})$. When $\V{\mu}^{(l)}$ is achieved, the $\M{K}^{(l)}$ can be obtained by minimizing the part of the negative log posterior \eqref{eq:nlp} depending only on $\M{K}_l$, 
	\begin{equation*}
	\begin{split}
	g(\M{K}_l) &= -\sum_{m=1}^M \log p(\V{y}^{(m,l)}|\V{\mu}^{(l)}, \M{K}_l)  - \log p(\M{K}_l|\M{K}_0) \\
	&= (M+\nu_0+2n-1)\log|\M{K}_l| + \sum_{m=1}^M (\V{y}^{(m,l)})^T \M{W}_{\V{\mu}^{(l)}} \M{K}_l^{-1} \M{W}_{\V{\mu}^{(l)}}^T \V{y}^{(m,l)} + tr(\M{K}_0 \M{K}_l^{-1}) \\
	&= (M+\nu_0+2n-1)\log|\M{K}_l| + tr\left[\left(\sum_{m=1}^M \M{W}_{\V{\mu}^{(l)}}^T \V{y}^{(m,l)}  (\V{y}^{(m,l)})^T \M{W}_{\V{\mu}^{(l)}} + \M{K}_0 \right) \M{K}_l^{-1} \right]
	\end{split}
	\end{equation*}
	The minimum is achieved at 
	\begin{equation} \label{eq:K}
	\M{K}_{l, j+1} = \frac{1}{M+\nu_0+2n-1} \left(\sum_{m=1}^M \M{W}_{\V{\mu}^{(l, j+1)}}^T \V{y}^{(m,l)}  (\V{y}^{(m,l)})^T \M{W}_{\V{\mu}^{(l, j+1)}} + \M{K}_0 \right).
	\end{equation}
	For the computational efficiency, one can run only one step of optimizing $\V{\mu}^{l}$ for each iteration. For checking the convergence, we check the magnitude of $||\M{W}_{\V{\mu}^{(l,j)}}^T \V{\mu}^{(l, j+1)}||_F$, which should be very small at the convergence.
	
	\section{Additional Two Use Cases} \label{EC3}
	This e-companion section demonstrates two use cases of the proposed motion analysis method.
	
	\subsection{Use Case 4: Analysis of Motion Variations} \label{sec:use3}
	We present the method in Section \ref{sec:motion} to analyze the variations of motions with the intention of understanding the overall mean of the worker's motions and the variation of other workers motions around the average. This is very useful to study and quantify the uncertainties in manual works within smart factories. For the motion analysis, we normalized out the rate information from the sixty posture sequences to achieve the rate-normalized posture sequences, $\tilde{\alpha}_m(t) = \alpha_m \circ \gamma_m(t)$. We applied Algorithm \ref{alg:MAP} to fit a probability distribution of motions to the sixty rate-normalized posture sequences. The algorithm gives the distribution parameters, $\V{\mu}^{(l)}$ and $\M{\Sigma}_l$, for the postures observed at each time $t_l$. 
	
\begin{figure}[t]
		\centering
		\includegraphics[width=\textwidth]{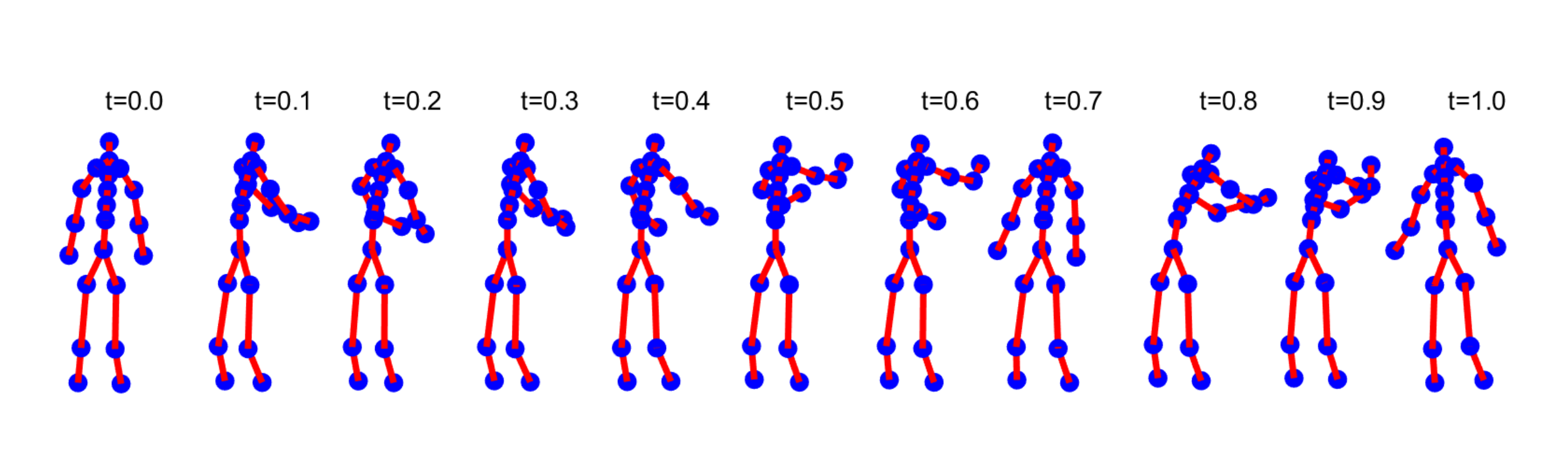}
		\caption{Posture Sequence of the Average Worker}
		\label{fig9}
	\end{figure}

	The mean postures $\V{\mu}^{(l)}$ are temporally correlated, and Figure \ref{fig9} shows the time-correlated mean postures over time, which can be interpreted as how the average worker performs the given work.   
	
\begin{figure}[t]
		\centering
		\includegraphics[width=\textwidth]{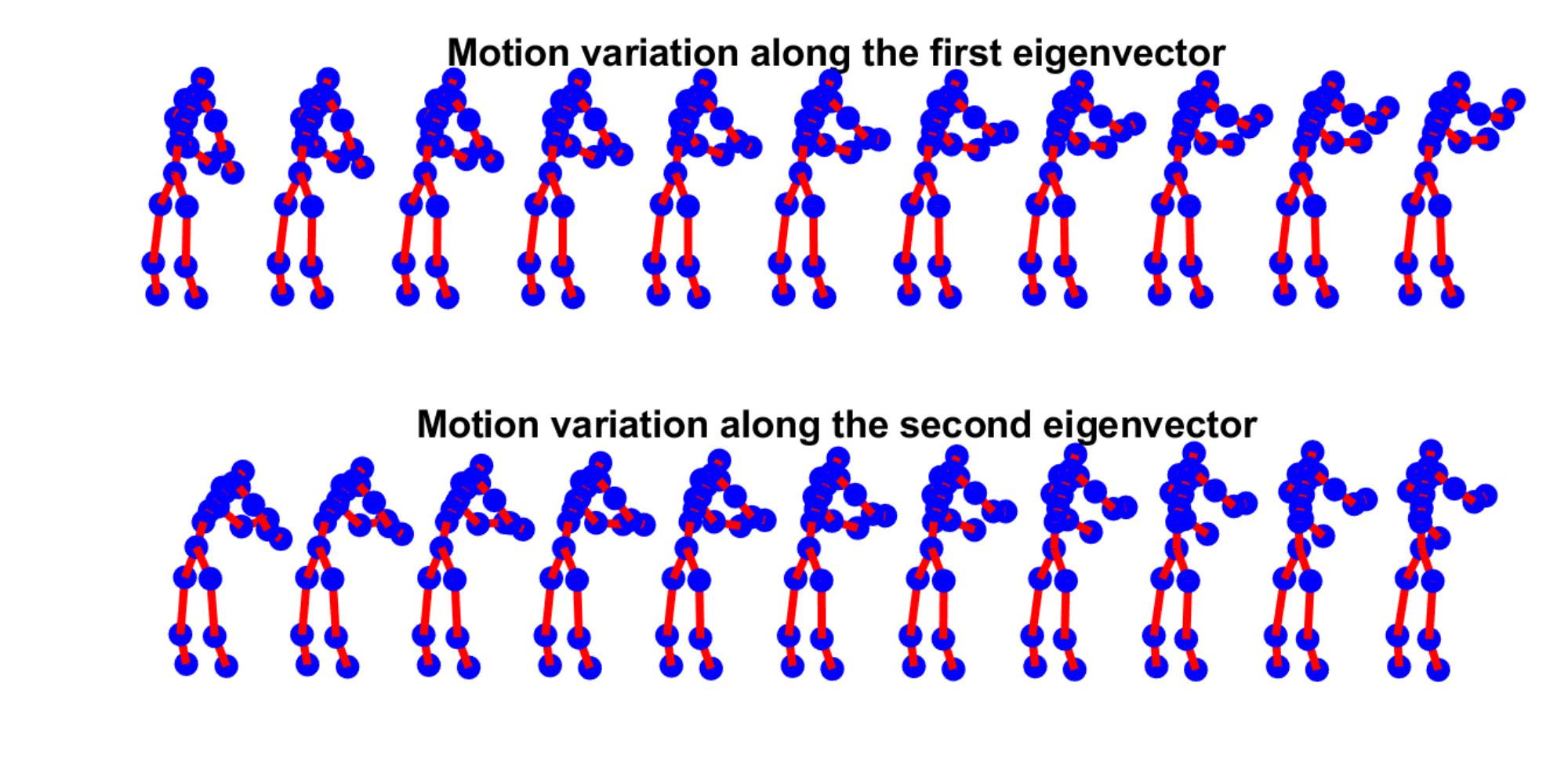}
		\caption{Motion variation of the work element performed at $t = 0.8$ along the first two eigenvectors of the motion covariance $\M{\Sigma}_l$.}
		\label{fig10}
	\end{figure}

	For each time point $t_l$, we can also analyze the variation of motions around the mean. Given the mean posture $\M{\mu}^{(l)}$ at $t_l$, the tangent coordinate of the posture observed at time $t_l$ has covariance matrix $\M{\Sigma}_l$. We can analyze the major eigenvectors of the covariance matrix to understand the major variation of the tangent coordinate and corresponding variation in the posture. Let $\V{v}_{\V{\mu}_l}$ denote the first eigenvector of the covariance matrix $\M{\Sigma}_l$. The variation of the tangent coordinate along the first eigenvector is described as the constant multiple of the eigenvector, 
	\begin{equation*}
	s \V{v}_{\V{\mu}_l},
	\end{equation*} 
	where $s \in \mathbb{R}$ is a constant. Varying the values of $s$, we can generate different tangent coordinates, each of which can be mapped to the corresponding posture by taking the exponential map, 
	\begin{equation*}
	\exp_{\V{mu}_l} (s \V{v}_{\V{\mu}_l}).
	\end{equation*}
	Figure \ref{fig10} shows the motion variation of the work performed at $t=0.8$ along the first two eigenvectors while $s$ ranges $[-1, 1]$. The first two eigenvectors explains 72\% of the total motion variation. 
	
	\subsection{Use Case 5: Work Restandardization} \label{sec:use4}
\begin{figure}[t]
		\centering
		\includegraphics[width=\textwidth]{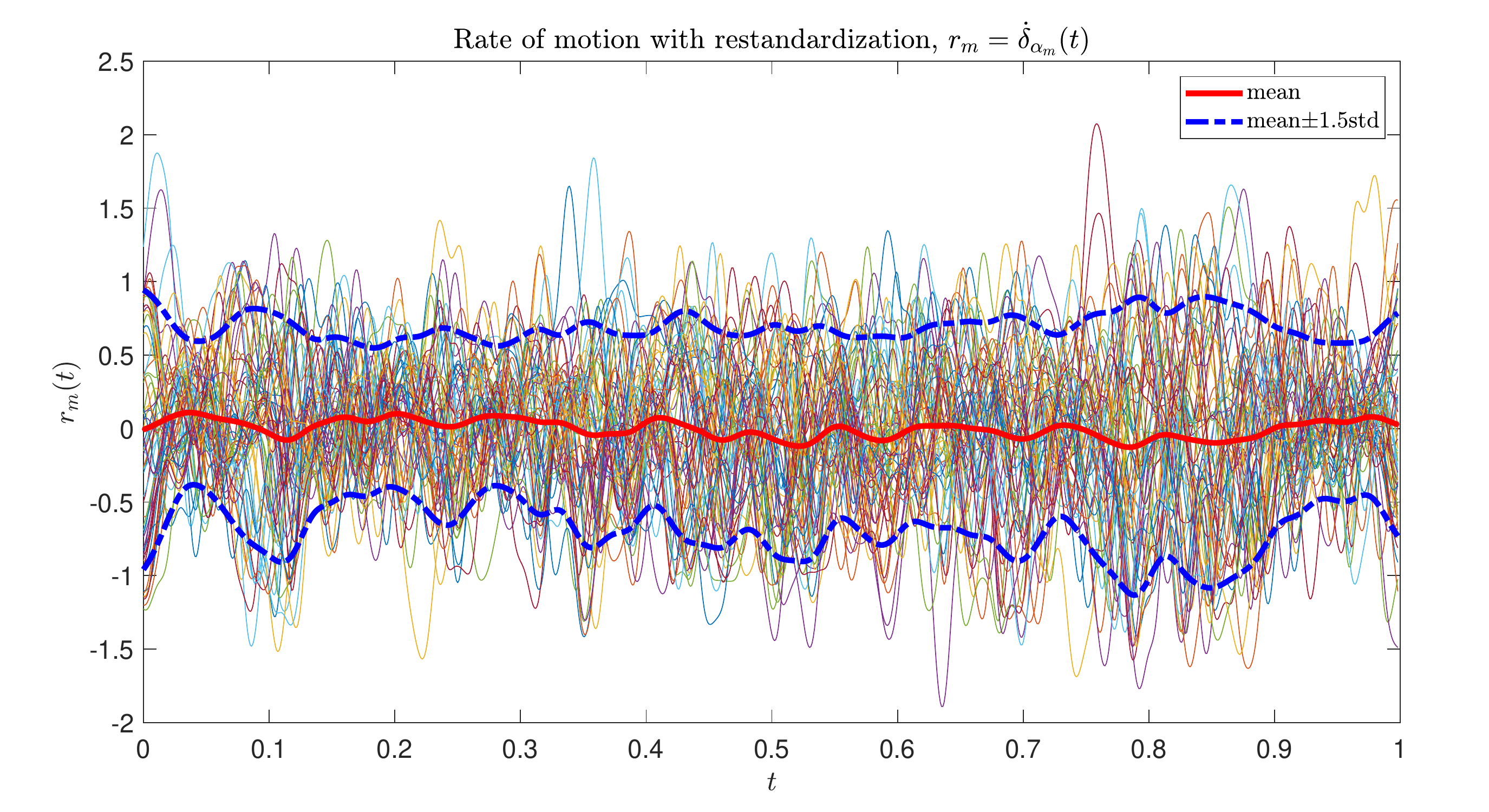}
		\caption{Time alignment functions after the work restandardization. }
		\label{fig7}
	\end{figure}

	As illustrated in Section \ref{sec:use2} and Figure \ref{fig4}, the analysis of variations in the rate functions revealed the temporal variability of motions. Figure \ref{fig4} shows the probability distribution of the rate functions fitted to the 60 rate functions displayed in the figure, following the Gaussian process model described in Section \ref{sec:rate_dist}. In the figure, the thick center line represents the mean rate function of the distribution, and the two other thick lines above and below the center line represents the 1.5 times standard deviations from the mean. The dotted line going through the zero base line corresponds to the rate function of the reference posture sequence. We can see the mean rate function significantly below the zero base line, which implies that most of the workers underperformed the standard operating procedure (SOP) represented by the reference posture sequence and the work rate described in the SOP is too fast. An operation manager can exploit this finding to redefine the SOP by adjusting the work rate more more reflecting the actual work paces. The adjustment can be achieved by taking the mean rate function $\bar{r}(t)$ and the corresponding time-alignment function $\bar{\gamma}(t) = \int_0^t \exp(\bar{r}(t')) dt' / \int_0^1 \exp(\bar{r}(t')) dt'$ and then by updating the reference posture sequence $\alpha_R$ with $\alpha_R \circ \bar{\gamma}$. After the adjustment, the time alignment functions can be recalculated with the temporally adjusted reference posture sequence, and the rate functions can be recalculated accordingly. Figure \ref{fig7} shows the rate functions after the adjustment. Now, the rate functions are centered around the zero base line, and the adjusted standard would reflect the actual work paces closely.  
	
\end{document}